\documentclass{article} % For LaTeX2e
\usepackage{iclr2018_conference,times}
\PassOptionsToPackage{hyphens}{url}\usepackage{hyperref}
\usepackage[hyphens]{url}
\usepackage[colorinlistoftodos,textsize=tiny]{todonotes}

\usepackage{graphicx} % more modern
\usepackage{subfigure} 
\usepackage{bm}
\usepackage{bbm}
\usepackage{amssymb,amsthm}
\usepackage[cmex10]{amsmath}
\usepackage[compact]{titlesec}
\usepackage{natbib}
\usepackage{multirow}
\usepackage{array}
\newcolumntype{L}[1]{>{\raggedright\let\newline\\\arraybackslash\hspace{0pt}}m{#1}}
\newcolumntype{C}[1]{>{\centering\let\newline\\\arraybackslash\hspace{0pt}}m{#1}}
\newcolumntype{R}[1]{>{\raggedleft\let\newline\\\arraybackslash\hspace{0pt}}m{#1}}

\usepackage{algorithm}
\usepackage{algorithmic}

\title{Latent Constraints: \\ Learning to Generate Conditionally from \\ Unconditional Generative Models}

\author{Jesse Engel \\ Google Brain \\ San Francisco, CA, USA
\And Matthew D. Hoffman \\ Google Inc. \\ San Francisco, CA, USA
\And Adam Roberts \\ Google Brain \\ San Francisco, CA, USA}
\newcommand{\Loss}{\mathcal{L}}
\newcommand{\E}{\operatorname{\mathbb E}}
\newcommand{\R}{\operatorname{\mathbb R}}

\newcommand{\CMaj}{\mathrm{C}_\mathrm{Maj}}
\newcommand{\mse}{\mathrm{MSE}}

\allowdisplaybreaks[2]
\graphicspath{{}}

\iclrfinalcopy % Uncomment forcamera-ready version

\begin{document}

\suppressfloats % no floats on title page

\maketitle

\begin{abstract}
Deep generative neural networks have proven effective at both conditional and unconditional modeling of complex data distributions.
Conditional generation enables interactive control, but creating new controls often requires expensive retraining. 
In this paper, we develop a method to condition generation without retraining the model. 
By post-hoc learning \emph{latent constraints}, value functions that identify regions in latent space that generate outputs with desired attributes, we can conditionally sample from these regions with gradient-based optimization or amortized actor functions.
Combining attribute constraints with a universal ``realism'' constraint, which enforces similarity to the data distribution, we generate realistic conditional images from an unconditional variational autoencoder.
Further, using gradient-based optimization, we demonstrate identity-preserving transformations that make
the minimal adjustment in latent space to modify the attributes of an image.
Finally, with discrete sequences of musical notes, we demonstrate zero-shot conditional generation, learning latent constraints in the absence of labeled data or a differentiable reward function. Code with dedicated cloud instance has been made publicly available (\href{https://goo.gl/STGMGx}{https://goo.gl/STGMGx}).

\end{abstract}

%%% ==============================================
\section{Introduction}
Generative modeling of complicated data such as images and audio is a long-standing challenge in machine learning.
While unconditional sampling is an interesting technical problem,
it is arguably of limited practical interest in its own right:
if one needs a non-specific image (or sound, song, document, etc.), one can
simply pull something at random from the unfathomably vast media databases on the web.
But that naive approach may not work for \emph{conditional} sampling (i.e., generating data to
match a set of user-specified attributes), since as more attributes are specified, it becomes exponentially less likely that a satisfactory example can be pulled from a database.
One might also want to \emph{modify} some attributes of an object while preserving its
core identity. These are crucial tasks in creative applications, where the typical user desires fine-grained controls \citep[][]{Bernardo2017}.

One can enforce user-specified constraints at training time, either by training on a curated subset of data or with conditioning variables. These approaches can be effective if there is enough labeled data available,
but they require expensive model retraining for each new set of constraints and may not leverage commonalities between tasks. Deep latent-variable models, such as Generative Adversarial Networks \citep[GANs; ][]{gan} and Variational Autoencoders \citep[VAEs; ][]{Kingma2013,rezende2014stochastic}, learn to unconditionally generate realistic and varied outputs by sampling from a semantically structured latent space. One might hope to leverage that structure in creating new conditional controls for sampling and transformations \citep[][]{Brock2016}.

Here, we show that new constraints can be enforced post-hoc on pre-trained unsupervised generative models. This approach removes the need to retrain the model for each new set of constraints, allowing users to more easily define custom behavior. We separate the problem into (1) creating an
%potentially large
unsupervised model that learns how to reconstruct data from latent embeddings, and (2) leveraging the latent structure exposed in that embedding space as a source of prior knowledge, upon which we can
%more easily
impose behavioral constraints.

Our key contributions are as follows:

\begin{itemize}
\item We show that it is possible to generate conditionally from an unconditional model, learning a critic function $D(z)$ in latent space and generating high-value samples with either gradient-based optimization or an amortized actor function $G(z)$, even with a non-differentiable decoder (e.g., discrete sequences).
\item Focusing on VAEs, we address the tradeoff between reconstruction quality and sample quality (without sacrificing diversity) by enforcing a universal ``realism'' constraint that requires samples in latent space to be indistinguishable from encoded data (rather than prior samples). 
\item Because we start from a VAE that can reconstruct inputs well,
we are able to apply \emph{identity-preserving transformations} by making
the minimal adjustment in latent space needed to satisfy the desired constraints.
For example, when we adjust a person's expression or hair, the result is still
clearly identifiable as the same person (see Figure \ref{fig:transform}).
This contrasts with pure GAN-based transformation approaches, which often fail to preserve identity.
\item \textit{Zero-shot conditional generation}. Using samples from the VAE to generate exemplars, we can learn an actor-critic pair that satisfies user-specified rule-based constraints in the absence of any labeled data.
\end{itemize}

%%% ==============================================
\begin{figure}[t]
\centering
\includegraphics[width=\textwidth]{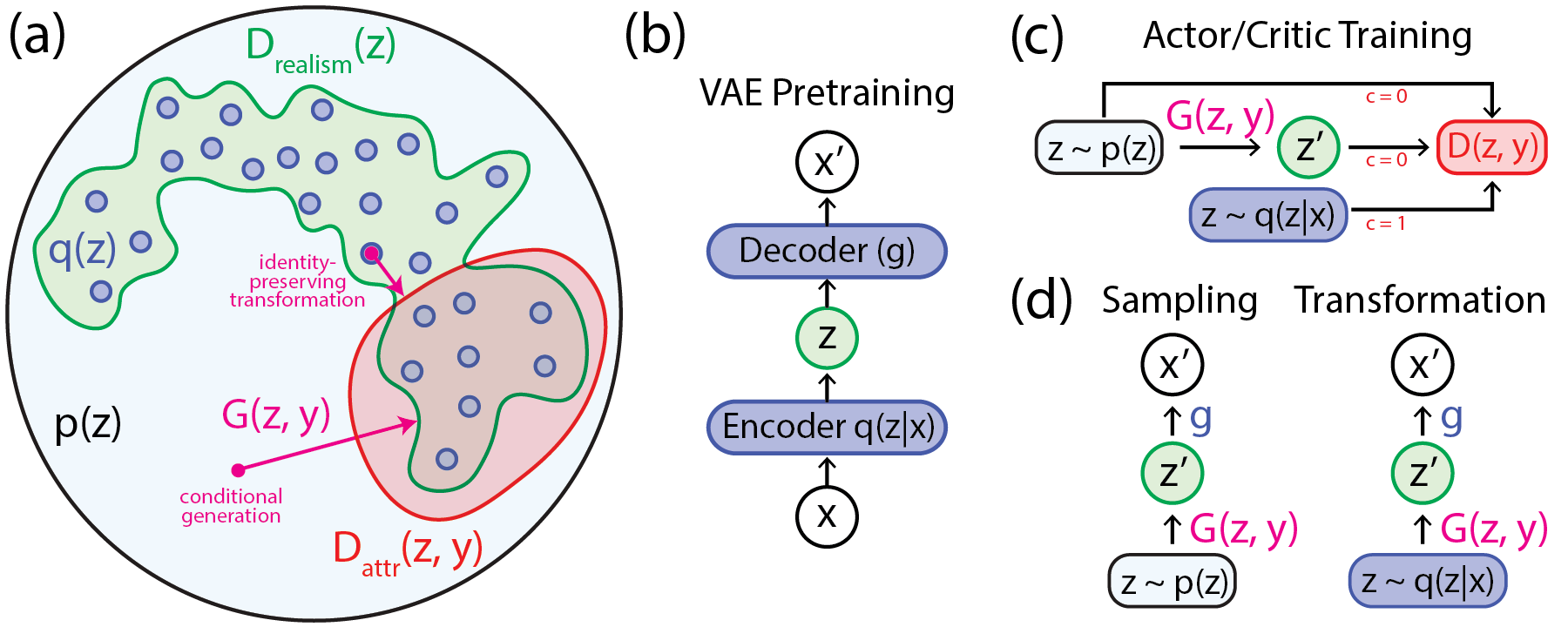}
\caption{(a) Diagram of latent constraints for a VAE. We use one critic $D_\mathrm{attr}$ to predict which
%distributions over
regions of the latent space will generate outputs with desired attributes,  and another critic $D_\mathrm{realism}$ to predict which regions have high mass under the marginal posterior, $q(z)$, of the training data. 
(b) We begin by pretraining a standard VAE, with an emphasis on achieving good reconstructions. 
(c) To train the actor-critic pair we use constraint-satisfaction labels, $c$, to train $D$ to discriminate between encodings of actual data, $z \sim q(z|x)$, versus latent vectors $z\sim p(z)$ sampled from the prior or transformed prior samples $G(z \sim p(z), y)$. Similar to a Conditional GAN, both $G$ and $D$ operate on a concatenation of $z$ and a binary attribute vector, $y$, allowing $G$ to learn conditional mappings in latent space. If $G$ is an optimizer, a separate attribute discriminator, $D_\mathrm{attr}$ is trained and the latent vector is optimized to reduce the cost of both $D_\mathrm{attr}$ and $D_\mathrm{realism}$.
(d) To sample from the intersection of these regions, we use either gradient-based optimization or an amortized generator, $G$, to shift latent samples from either the prior ($z \sim p(z)$, sampling) or from the data ($z \sim q(z|x)$, transformation). }
\label{fig:diagram}
\end{figure}

% %%% ==============================================
% \begin{table}[ht]
% \centering
% \begin{tabular}{ l | c c c c c }
%   Section & Critic Loss & Actor Loss &  \\
%  \hline

%                         & $\E_{z \sim q(z \mid x)}[\Loss_{c=1}(z)] + $ & $\E_{z \sim p(z)}[\Loss_{c=1}(G(z)) +$\\ 
% Conditional Generation  & $\E_{z \sim p(z)}[\Loss_{c=0}(z)] + $ &    $\lambda_\mathrm{dist} \Loss_{\mathrm{dist}}(G(z), z)]$\\
%             & $\E_{z \sim G(p(z))}[\Loss_{c=0}(z)]$ \\

%  Transformations & - & -  \\
%  Zero-Shot &  & 0.903 &  \\
% \end{tabular}
% \caption{Summary of cost functions used in each section.}
% \end{table}
% %%% ==============================================

%%% ==============================================
\section{Background}

%%% ==============================================

\begin{figure}[t]
\centering
\includegraphics[width=\textwidth]{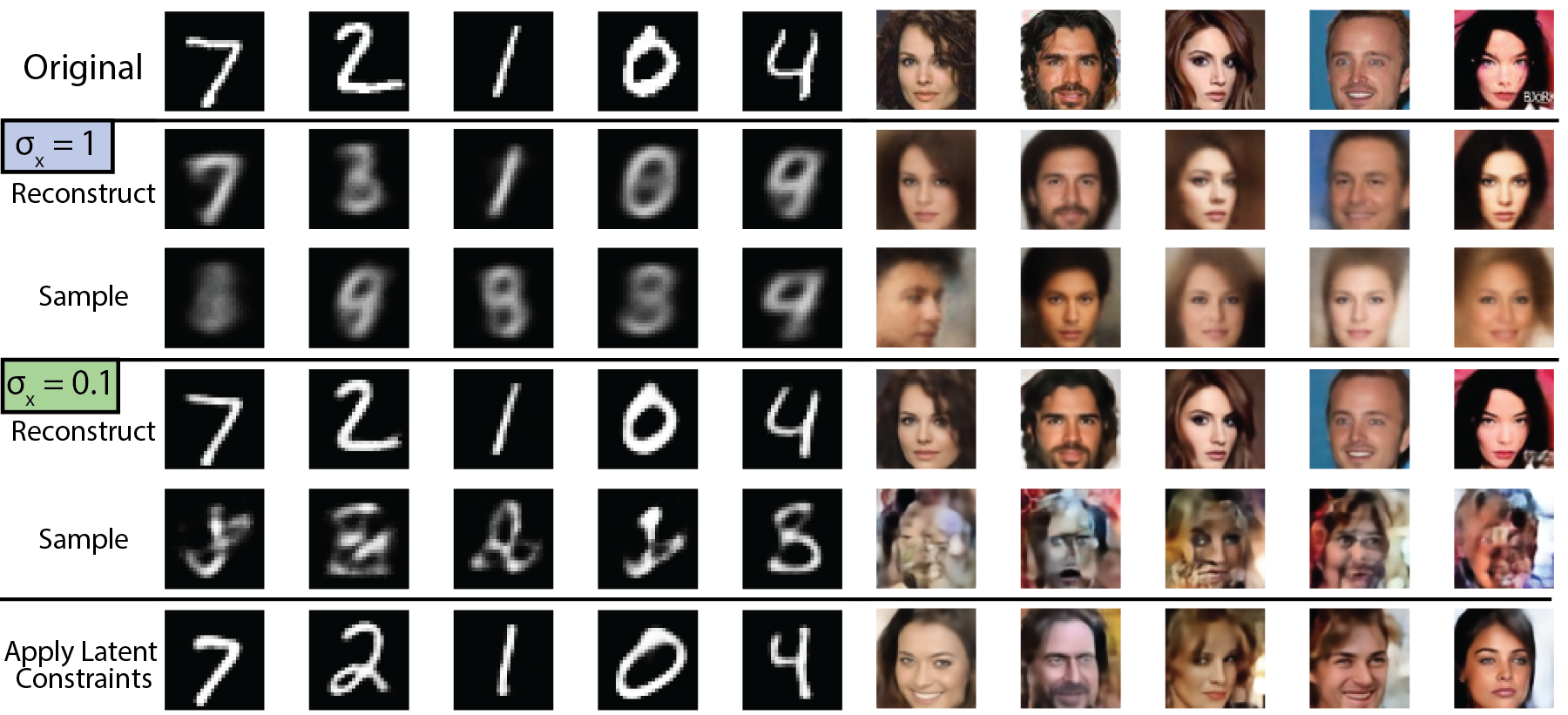}
\caption{Typical VAEs use a pixel-wise data likelihood, $\mathcal{N}(\mu_x(z), \sigma_x I)$,
with $\sigma_x=1$ to produce coherent samples at the expense of visual and conceptual blurriness (Row 3). Some reconstructions (Row 2) actually change attributes of the original data. Decreasing $\sigma_x$ to 0.1 maximizes the ELBO (supplemental Table~\ref{tab:ELBO}) and increases the fidelity of reconstructions (Row 4) at the cost of sample realism (Row 5). Using an actor to shift prior samples to satisfy the realism constraint, we achieve more realistic samples without sacrificing sharpness (Row 6). The samples are mapped to the closest point in latent space that both satisfies the realism constraint and has the same attributes as the original data.}
\label{fig:vae}
\end{figure}

%%% ==============================================

Decoder-based deep generative models such as VAEs and GANs generate samples that approximate a population distribution $p^\star(x)$ by passing samples from some
simple tractable distribution $p(z)$ (often $p(z)\triangleq\mathcal{N}(0, I)$)
through a deep neural network. GANs are trained to fool an auxiliary classifier that tries to learn
to distinguish between real and synthetic samples. VAEs are fit to data using a variational approximation to maximum-likelihood estimation:
\begin{equation}
\textstyle
\mathcal{L}^{\mathrm{ELBO}} \triangleq
\frac{1}{N}\sum_n \mathbb{E}_{z\sim q(z\mid x_n)}[\log \pi(x_n; g(z))]
- \mathrm{KL}(q(z\mid x_n)\mid\mid p(z)) 
\le \frac{1}{N}\sum_n \log p(x_n),
\end{equation}
where the ``encoder'' distribution $q(z\mid x)$ is an approximation to the posterior $p(z\mid x)$,
$\pi(x; g(z))\triangleq p(x\mid z)$ is a tractable likelihood function that depends on
some parameters output by a ``decoder'' function $g(z)$, and $q$ and $g$ are fit to maximize
the evidence lower bound (ELBO) $\mathcal{L}^\mathrm{ELBO}$. The likelihood $\pi(x;g)$ is often chosen to
be a product of simple distributions such as $\pi(x;g) = \mathcal{N}(x; g, \sigma_x^2 I)$ for
continuous data or $\pi(x;g) = \prod_d \mathrm{Bernoulli}(x_d; g_d)$ for binary data.

% are very effective at mapping points in a latent space to realistic outputs that exhibit various attributes.
% We can define an attribute as any feature that could be detected with a discriminative model or rule. 
% For example, generated images of handwritten digits can have varying line thicknesses and represent different numbers. 
% Generated faces can vary in attributes such as their age, hair color, and gender.
% In music, melodies can be played in different sets of pitches and different densities of notes. 

GANs and VAEs have complementary strengths and weaknesses.
GANs suffer from the ``mode-collapse'' problem, where the generator
assigns mass to a small subset of the support of the population
distribution---that is, it may generate realistic samples, but
there are many more realistic samples that it cannot generate.
This is particularly problematic if we want to use GANs to \emph{manipulate}
data rather than generate new data; even GAN variants that include some
kind of inference machinery \citep[e.g., ][]{donahue2016adversarial, Dumoulin2016, Perarnau2016}
to determine what $z$ best matches some $x$
tend to produce reconstructions that are reminiscent of the
input but do not preserve its identity.

On the other hand, VAEs (especially those with simple likelihoods $\pi$)
often exhibit a tradeoff between sharp reconstructions and sensible-looking samples (see Figure \ref{fig:vae}).
That is, depending on what hyperparameters they are trained with
(e.g., latent dimensionality and the scale of the likelihood term),
VAEs tend to either produce blurry reconstructions and plausible (but blurry) novel samples,
or bizarre samples but sharp reconstructions.
It has been argued \citep{AAE} that this is due to the ``holes'' problem; the decoder is trained on samples from the marginal posterior $q(z)\triangleq \frac{1}{N}\sum_n q(z\mid x_n)$, which may have very high KL divergence to the presupposed marginal $p(z)$ \citep{ELBOSurgery}. 
In particular, if the decoder, $g(z)$, can reconstruct arbitrary values of $x$ with high accuracy (as in the case of small $\sigma_x$) then the typical posterior $p(z\mid x)$ will be highly concentrated. We show this experimentally in supplemental Figure~\ref{fig:sigmamean}. If $q(z\mid x)$ underestimates the posterior variance (as it usually does),
then the marginal posterior $q(z)$ will also be highly concentrated, and samples from $p(x)=\int_z p(z)p(x\mid z)dz$ may produce results that are far from typical reconstructions $\mathbb{E}_p[x\mid z\sim q(z\mid x)]$. 
If we tune $\sigma_x$ to maximize the ELBO \citep{bishop2006}, we find the optimal $\sigma_x\approx 0.1$ (supplemental Table~\ref{tab:ELBO}). Figure~\ref{fig:vae} shows that this choice does indeed lead to good reconstructions but strange-looking samples.

Conditional GANs \citep[CGAN; ][]{Mirza2014} and conditional VAEs \citep[CVAE; ][]{Sohn2015} 
can generate samples conditioned on attribute information when available, but they 
must be trained with knowledge of the attribute labels for the whole
training set, and it is not clear how to adapt them to new attributes without retraining from scratch.
Furthermore, CGANs and CVAEs suffer from the same
problems of mode-collapse and blurriness as their unconditional cousins.

We take a different approach to conditional generation and identity-preserving
transformation. We begin by training an unconditional VAE with hyperparameters
chosen to ensure good reconstruction (at the expense of sample quality).
We then train a ``realism'' critic to predict whether a given $z$ maps to a
high-quality sample. We also train critics to predict whether a given
$z$ maps to a sample that manifests various attributes of interest.
To generate samples that are both realistic and exhibit desired attributes,
one option is to optimize random $z$ vectors until they
satisfy both the realism and attribute critics. Alternately, we can
amortize this cost by training an ``actor'' network to map a random set of $z$ vectors
to a subregion of latent space that satisfies the constraints encoded by the critics. By encouraging these
transformed $z$ vectors to remain as close as possible to where they started, we alleviate
the mode-collapse problem common to GANs.

% Given a constraint on the generated outputs $c(x) : \R^N \rightarrow [0, 1]$, we would like to learn a distribution $r(z)$ within the latent space, such that sampling  $z' \sim r(z)$ produces outputs $x' \sim p(x|z')$ that satisfy the constraint $c(x') = 1$.
% It can be helpful to think of the problem in terms of reinforcement learning, where $c(x)$ provides a reward function, which may or may not be differentiable. Our goal is then to learn a critic value function $D(z) : \R^M \rightarrow \[0, 1\]$ that approximates the true reward, and an actor function or procedure $G(z) : \R^M \rightarrow \R^M$ that can map samples from the prior with probability proportional to the value function $z' \sim r(z) \propto D(z) p(z)$.
% For simplicity, we use the symbols $G$ and $D$ to represent the actor and critic functions respectively, as a natural analog exists to Generative Adversarial Networks, where the generator maps from one part of latent space to another, and the discriminator guesses whether that part of latent space corresponds to outputs that satisfy the constraints. We refer to the critic function as a \emph{Latent Constraint}, as it moves the constraint satisfaction problem from the data space to the latent space. The actor is then a function or procedure that learns to satisfy the latent constraint. 

Our approach is summarized visually in Figure~\ref{fig:diagram}. The details
follow in sections \ref{sec:realism}, \ref{sec:attributes}, \ref{sec:identity},
and \ref{sec:rules}.

% As diagrammed in Figure~\ref{fig:diagram}, we take the approach of conditionally generating from such unconditional generative models by using supervised learning to identify regions of latent space that correspond to generated samples with desired attributes. 
% We can then generate samples from these regions using either gradient-based optimization or an amortized actor function.
% For VAEs, we find that we can sharpen the quality of generated samples by constraining generation to regions enclosing those of the training data. 
% Further, we perform transformations to existing data that preserve the identity of the data by shifting the minimmum amount in latent space to reach the appropriate subregion.
% Finally, we demonstrate that it is possible to learn to generate conditionally without any labeled training data, sampling from the VAE prior to generate exemplars whose value we evaluate with hard-coded rules. 
% While, for clarity, we will focus the remainder of the paper on VAEs, we make no specific assumptions that would prohibit these techniques from being applicable to other latent variable models.

\section{The ``Realism'' Constraint: Sharpening VAE Samples}
\label{sec:realism}

%%% ==============================================
\begin{figure}[t]
\includegraphics[width=\textwidth]{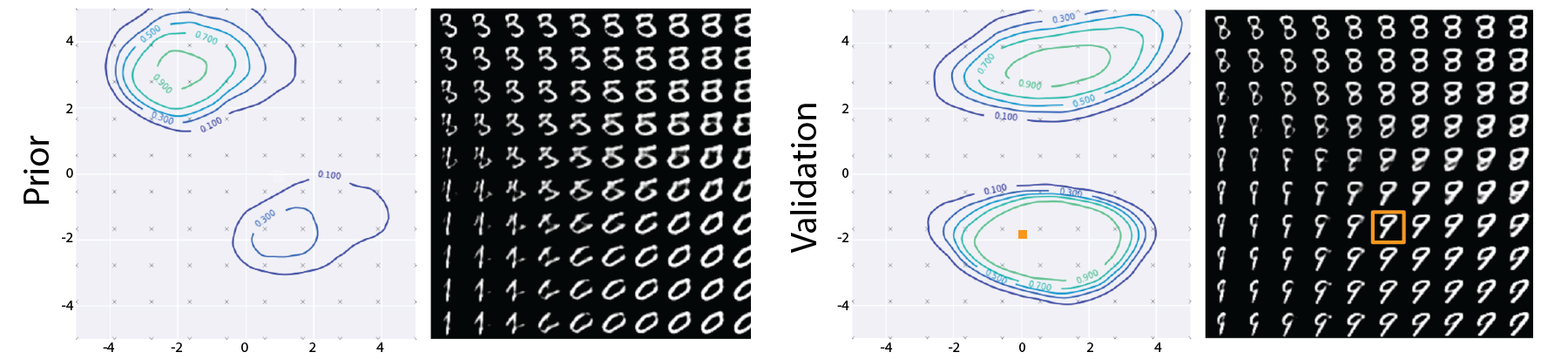}
\caption{Contour maps of the critic value functions for the marginal posterior (``realism") constraint. We look at the two latent dimensions that have the lowest average posterior standard deviation
on the training set, taking low variance in $z$ space as a proxy for influence over the generated images. All other latent dimensions are held fixed at their original values (from a sample from $p(z)$ on the left, and from a sample from $q(z\mid x)$ for a held-out $x$ on the right).
Gray x marks correspond to the points in latent space of the generated images to the right.
The cross-section on the left, taken from a prior sample, shows contours that point towards more realistic looking digits. In the cross-section on the right, a sample from the validation set (indicated by orange squares) resides within a local maximum of the critic, as one would hope.}
\label{fig:contours}
\end{figure}
%%% ==============================================

We define the realism constraint implicitly as being satisfied by samples from the marginal posterior $q(z)\triangleq\frac{1}{N}\sum_n q(z\mid x_n)$ and not those from $p(z)$. By enforcing this constraint, we can close the gap between reconstruction quality and sample quality (without sacrificing sample diversity).

As shown in Figure~\ref{fig:diagram}, we can train a critic $D$ to differentiate between samples from $p(z)$ and $q(z)$.
% mhoffman: I'm not sure the likelihood-ratio trick name-check clarifies much here, actually.
%           The distribution we wind up sampling from isn't that easy to define.
% This equates to a VAE that performs implicit variational inference using the likelihood ratio trick \citep{alphaGAN}. 
%TODO(deck): Can you give a line saying what the "likelihood ratio trick" is?
% Unlike previous methods that compare $p(z)$ to another known distribution, the realism constraint compares to the marginal posterior $q(z)$ of the pretrained model.
The critic loss, $\Loss_D(z)$, is simply the cross-entropy, with labels $c=1$ for $z \sim q(z\mid x)$ and $c=0$ for $z \sim p(z)$. We found that the realism critic had little trouble generalizing to unseen data;
that is, it was able to recognize samples from $q(z\mid x^\mathrm{held-out})$ as being ``realistic'' (Figure~\ref{fig:contours}).
%TODO(mhoffman): Latex hack: don't put blank lines around equations. It wastes a ton of space. Putting a % on the line solves this problem.
%TODO(mhoffman): Another latex note: Always prefer $\log$ to $log$. More generally, when you want to say that something is a word as opposed to a product of variables, it's good to wrap the word in \mathrm{}. IMO subscripts like \lambda_{penalty} look better as \lambda_\mathrm{penalty}.
% \begin{equation}\begin{split}
% &\Loss_D(z) = \E_{z \sim q(z \mid x)}[-\log(D(z))] + \E_{z \sim p(z)}[-\log(1-D(z))] 
% \end{split}\end{equation}
% %

%TODO(mhoffman): Add reference on hard-negative mining.
Sampling from the prior is sufficient to train $D$ for models with lower KL Divergence, but if the KL Divergence between $q$ and $p$ is large, the chances of sampling a point $p(z)$ that has high probability under $q(z)$ becomes vanishingly small. This leads to poor sample quality and makes it difficult for $D$ to learn a tight approximation of $q(z)$ solely by sampling from $p(z)$. Instead, we use an inner-loop of gradient-based optimization, $G_{\mathrm{opt}}(z) = \mathrm{GradientDescent}(z; \Loss_D(z))$, to move prior samples to points deemed more like $q(z)$ by $D$. For clarity, we introduce the shorthand $\Loss_{c=1}(z) \triangleq -\log(D(z))$ and $\Loss_{c=0}(z) \triangleq -(1 - \log(D(z)))$. This gives us our critic loss for the realism constraint:
\begin{equation}\begin{split}
&\Loss_D(z) = \E_{z \sim q(z \mid x)}[\Loss_{c=1}(z)] + \E_{z \sim p(z)}[\Loss_{c=0}(z)] + \E_{z \sim G(p(z))}[\Loss_{c=0}(z)]
\end{split}\label{eq:critic}\end{equation}
Since this inner-loop of optimization can slow down training, we amortize the generation by using a neural network as a function approximator. There are many examples of such amortization tricks, including the encoder of a VAE, generator of a GAN, and fast neural style transfer \citep{Ulyanov2016,li2016precomputed,johnson2016perceptual}. As with a traditional GAN, the parameters of the function $G$ are updated to maximize the value $D$ ascribes to the shifted latent points. One of the challenges using a GAN in this situation is that it is prone to mode-collapse. However, an advantage of applying the GAN in latent space is that we can regularize $G$ to try and find the closest point in latent space that satisfies $D$, thus encouraging diverse solutions. We introduce a regularization term, $\Loss_{\mathrm{dist}}(z', z) = 1/\bar{\sigma_z}^2 \log(1 + (z'- z)^2)$ to encourage nearby solutions, while allowing more exploration than a mean square error term. As a VAE utilizes only a fraction of its latent dimensions, we scale the distance penalty of each dimension by its utilization, as indicated by the squared reciprocal of the scale $\sigma_z(x)$ of the encoder distribution $q(z\mid x)$, averaged over the training dataset, $\bar{\sigma}_z \triangleq \frac{1}{N}\sum_n \sigma_z(x_n)$.
The regularized loss is
\begin{equation}\begin{split}
&\Loss_G(z) = \E_{z \sim p(z)}[\Loss_{c=1}(G(z)) + \lambda_\mathrm{dist} \Loss_{\mathrm{dist}}(G(z), z)].
\end{split}\label{eq:actor}\end{equation}
%

% MATT: Describe motivation and details of the realism constraint. 
% Motivate the use of x_sigma = 0.1, Figure~\ref{fig:vae} 
% Also motivate having lots of latent dimensions (1024 were used for all VAEs, except melody (512))

%%% ========================    ======================
\begin{figure}[t]
\centering
\includegraphics[width=0.9\textwidth]{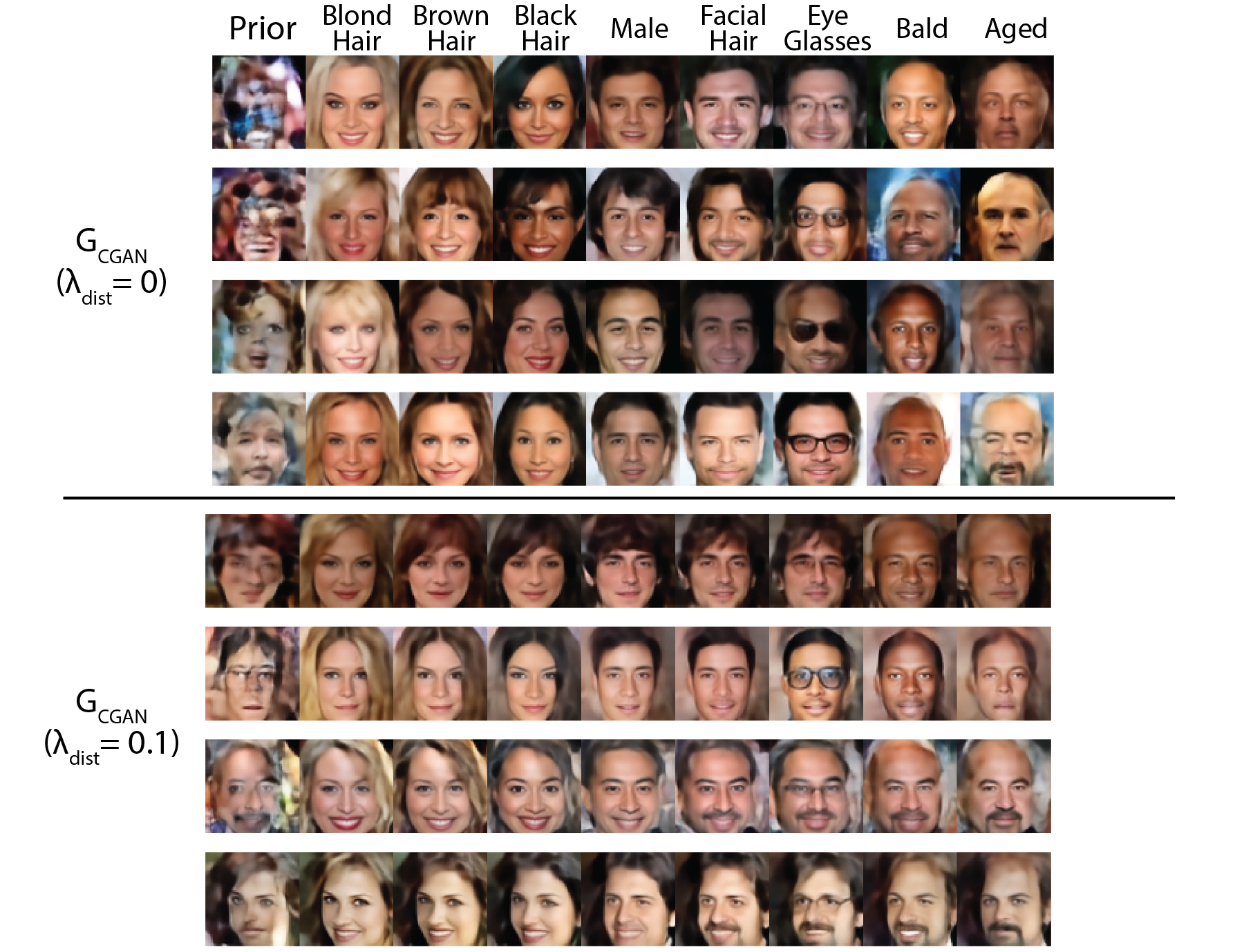}
\caption{Conditional generation with a CGAN actor-critic pair acting in the latent space of a VAE with $\sigma_x = 0.1$. Each row starts from a different prior sample and maps it to a new point in latent space that satisfies both the attribute constraints and the realism constraint. The attribute constraints are changed one at a time to produce as smooth a transition as possible from left to right. The bottom CGAN is regularized during training to prefer small shifts in latent space ($\lambda_\mathrm{dist} = 0.1$),  while the top is not ($\lambda_\mathrm{dist} = 0.0$).
Compared to the images generated by the unregularized model, the images generated by the regularized model are much less diverse across columns, suggesting that the regularization does indeed enforce some degree of identity preservation.
The regularized model produces images that are somewhat \emph{more} diverse across rows, suggesting that the regularization fights mode collapse (arguably at the expense of image quality).
% fights mode-collapse, leading to local solutions that are strongly similar for each row, arguably at the expense of image quality.
For each column, the complete list of attributes is given in supplemental~Table~\ref{tab:attrlist}.}
\label{fig:cgan}
\end{figure}
%%% ==============================================

%%% ==============================================
\begin{figure}[ht]
\centering
\includegraphics[width=0.9\textwidth]{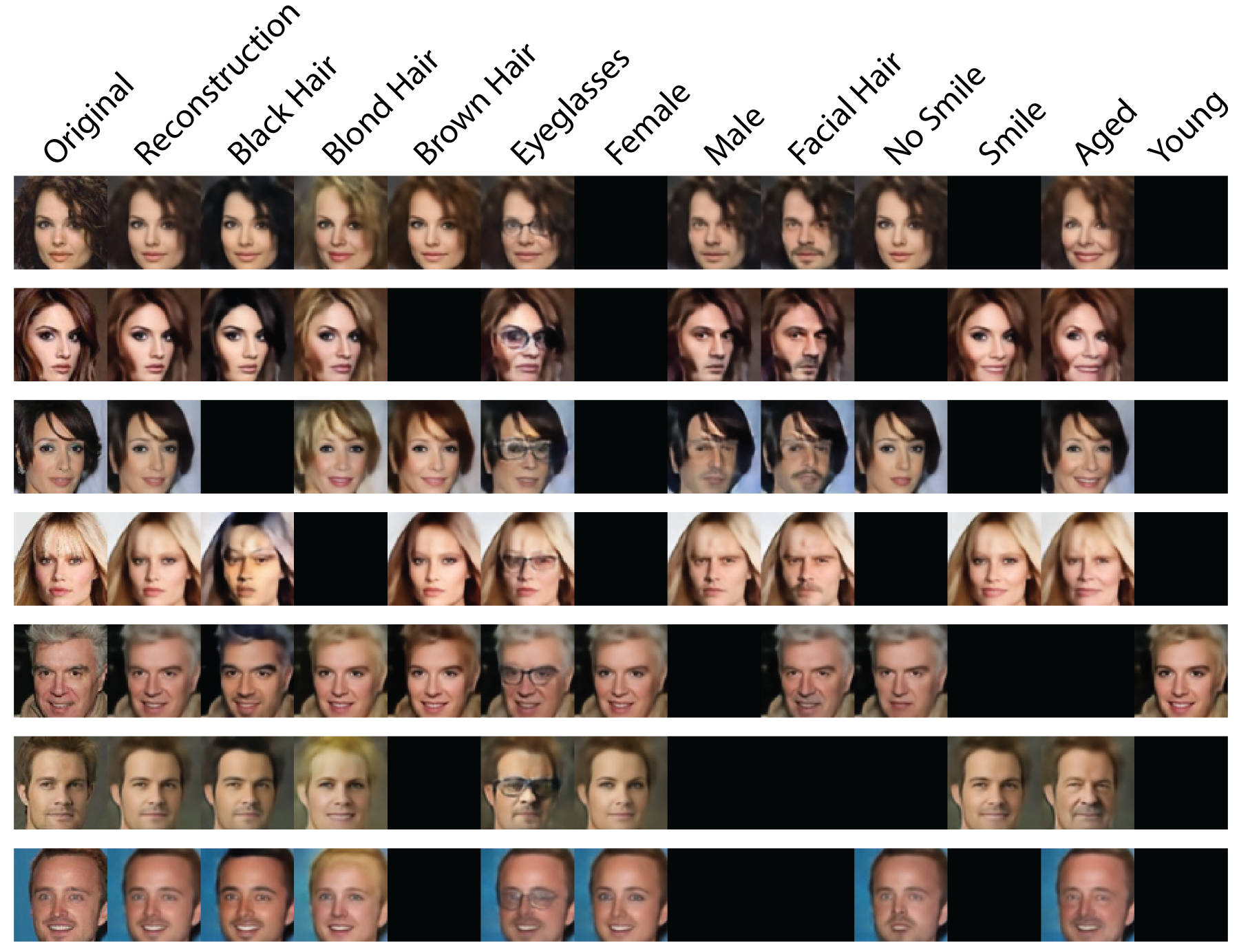}
\caption{Identity-preserving transformations with optimization. Two separate critics are trained, one for attributes and one for the realism constraint. Starting at the latent points corresponding to the data reconstructions, we then perform gradient ascent in latent space on a weighted combination of critic values (1.0 attribute, 0.1 marginal posterior), stopping when a threshold value is passed for both critics. Images remain semantically close to the original because the pixel-wise likelihood of VAE training encourages identity-preserving reconstructions, and the dynamics of gradient ascent are naturally limited to finding solutions close in latent space. Panels are black for attributes of the original image, as the procedure just returns the original point in latent space.}
\label{fig:transform}
\end{figure}
%%% ==============================================

%%% ==============================================
\section{Attribute Constraints: Conditional Generation}
\label{sec:attributes}

%%% ==============================================
\begin{table}[ht]
\centering
\begin{tabular}{ l | c c c c c }
  CelebA & Accuracy & Precision & Recall & F1 Score & \textbf{$z_\mse$}  \\
 \hline
 (This Work) 10 Attributes \\
 Test Data & 0.936 & 0.901 & 0.893 & 0.895 &  \\ 
 $G_{CGAN} (\lambda_\mathrm{dist} = 0)$ & 0.942 & 0.914 & 0.904 & 0.906 & 80.7 \\
 $G_{CGAN} (\lambda_\mathrm{dist} = 0)$ (Small Model) & 0.926 & 0.898 & 0.860 & 0.870 & 58.9 \\
 $G_{CGAN} (\lambda_\mathrm{dist} = 0.1)$ & 0.928 & 0.903 & 0.863 & 0.874 & 17.0 \\
 \hline
 \citep[][]{Perarnau2016} 18 Attributes \\
 Test Data& 0.928 &  &  & 0.715 &  \\
 IcGAN & 0.860 &  &  & 0.524 &  \\
\end{tabular}
\caption{Accuracy of a separate model trained to classify attributes from images, evaluated on test data and generated images. We condition and evaluate the generated images on the same labels as the test data. For comparison, the results of a similar task using invertible CGANs for generation \citep[][]{Perarnau2016} are provided. However, since the full list of salient attributes was not given in the paper, we emphasize that they are not directly comparable as the two experiments use a slightly different set of attribute labels. We also measure the distance in latent space that prior samples are shifted, weighted by $1/\bar{\sigma_z}^2$. Actors trained with a latent distance penalty $\lambda_\mathrm{dist}$ have slightly worse accuracy, but find latent points much closer to the prior samples and produce a greater diversity of images (see supplemental Figures~\ref{fig:suppcgan}~and~\ref{fig:suppcgandist}). Interestingly, an actor trained without a distance penalty achieves higher classification accuracy than the test set itself, possibly by generating images with more exaggerated and distinctive features than real data. A "small model" CGAN with 85x fewer parameters (3 fully connected layers of 256 units) generates images (supplemental Figure~\ref{fig:smallmodel}) of comperable quality. Due to the smaller capacity, the model finds more local solutions (smaller $z_{MSE}$) that have slightly less attribute accuracy, but are more visually similar to the prior sample without an explicit regularization term.}
\label{tab:condgen}
\end{table}
%%% ==============================================

We want to generate samples that are realistic, but we also want to control what attributes they exhibit. Given binary attribute labels $y$ for a dataset, we can accomplish this by using a CGAN in the latent space, which amounts to replacing $D(z)$ and $G(z)$ with conditional versions $D(z, y)$  and $G(z, y)$ and concatenating $y$ to $z$ as input. If both the actor and critic see attribute information, $G$ must find points in latent space that could be samples from $q(z)$ with attributes $y$. 

\begin{figure}[ht]
\includegraphics[width=\textwidth]{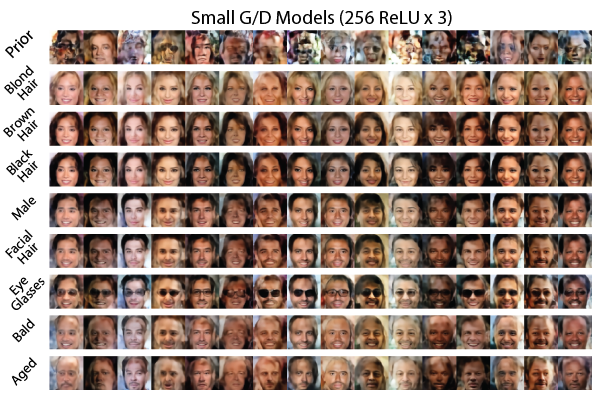}
\caption{Samples generated with smaller (3 ReLU layers of 256 units each) $G$ and $D$ models are comparable quality despite having 85x fewer parameters, $\lambda_\mathrm{dist} = 0.0$. Full attribute labels are given in supplementary Table~\ref{tab:attrlist}.}
\label{fig:smallmodel}
\end{figure}

This procedure is computationally inexpensive relative to training a generative model from scratch. In most of our experiments, we use a relatively large CGAN actor-critic pair (4 fully connected ReLU layers of 2048 units each), which during training uses about $96\times$ fewer FLOPs/iteration than the unconditional VAE. We also trained a much smaller CGAN actor-critic pair (3 fully connected ReLU layers of 256 units), which uses about $2884\times$ fewer FLOPs/iteration than the VAE, and achieves only slightly worse results than the larger CGAN (supplemental Figure~\ref{fig:smallmodel} and Table~\ref{tab:condgen}).

Figure~\ref{fig:cgan} demonstrates the quality of conditional samples from a CGAN actor-critic pair and the effect of the distance penalty, which constrains generation to be closer to the prior sample, maintaining similarity between samples with different attributes. 
The regularized CGAN actor has less freedom to ignore modes by pushing many random $z$ vectors to the same area of the latent space, since it is penalized for moving samples from $p(z)$ too far.
The increased diversity across rows of the regularized CGAN is evidence that this regularization does fight mode-collapse (additional qualitative evidence is in supplemental Figures~\ref{fig:suppcgan}~and~\ref{fig:suppcgandist}).
% This is evidence of a lack of mode-collapse, as can also be seen qualitatively by the greater diversity of images from many prior samples in the Appendix (supplemental Figures~\ref{fig:suppcgan}~and~\ref{fig:suppcgandist}).
However, without a distance penalty, samples appear more a bit realistic with more prominent attributes. This is supported by Table~\ref{tab:condgen}, where we use a separately trained attribute classification model to quantitatively evaluate samples. The actor with no penalty generates samples that are more accurately classified than the actor with a penalty but also shifts the samples much farther in latent space.

Although we used a VAE as the base generative model, our approach could also be used to generate high-quality conditional samples from pretrained classical autoencoders. We show in supplemental Figure~\ref{fig:nokl} that we obtain reasonably good conditional samples (albeit with high-frequency spatial artifacts) as $\sigma_x \rightarrow 0$ (equivalent to a classical autoencoder). 
Learning the decoder using VAE training encourages q(z) to fill up as much of the latent space as possible (without sacrificing reconstruction quality), which in turn encourages the decoder to map more of the latent space to reasonable-looking images. The prior $p(z) = \mathcal{N}(0, I)$ also imposes a natural scale on the latent variables.

\section{Identity-Preserving Transformations}
\label{sec:identity}
\begin{figure}[ht]
\centering
\includegraphics[width=0.5\textwidth]{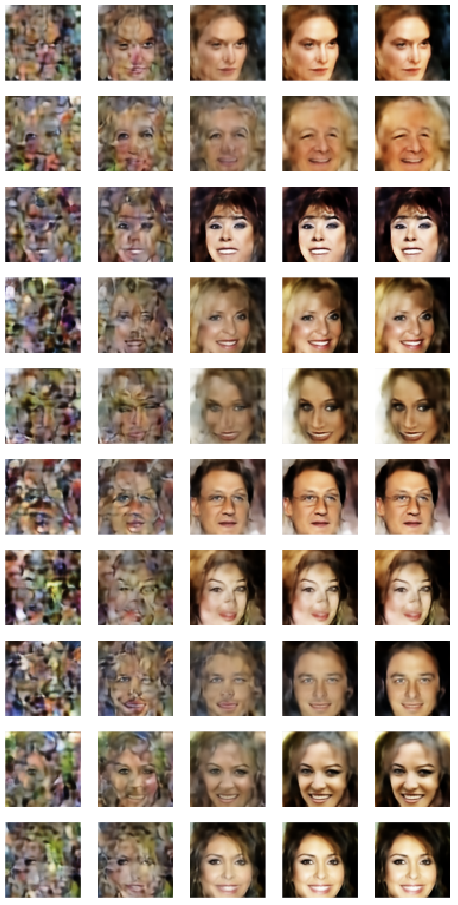}
\caption{Optimization of samples drawn from the prior to satisfy both the realism constraint and attribute constraints (drawn from the test set). The optimization takes 100 steps, and images are shown at 0, 10, 30, 50 and 100 steps.  $D$ is trained with inner-loop optimization, $G_{\mathrm{opt}}$, as described in Section~\ref{sec:arch}}
\label{fig:Opt}
\end{figure}

If we have a VAE that can produce good reconstructions of held-out data, we can transform the attributes of the output by gradient-based optimization. We simply need to train a critic, $D_{attr}(z)$, to predict the attribute labels $p(y\mid z)$ of the data embeddings $z \sim q(z\mid x)$, and use a cross-entropy loss to train. Then, starting from a data point, $z \sim q(z\mid x)$, we can perform gradient descent on the the realism constraint and attribute constraint jointly, $\Loss_{D_\mathrm{real}}(z) + \lambda_\mathrm{attr} \Loss_{D_\mathrm{attr}}(z)$. Note that it is helpful to maintain the realism constraint to keep the image from distorting unrealistically. Using the same procedure, we can also conditionally generate new samples (supplemental Figure~\ref{fig:Opt}) by starting from $z \sim p(z)$.

\begin{figure}[ht]
\includegraphics[width=\textwidth]{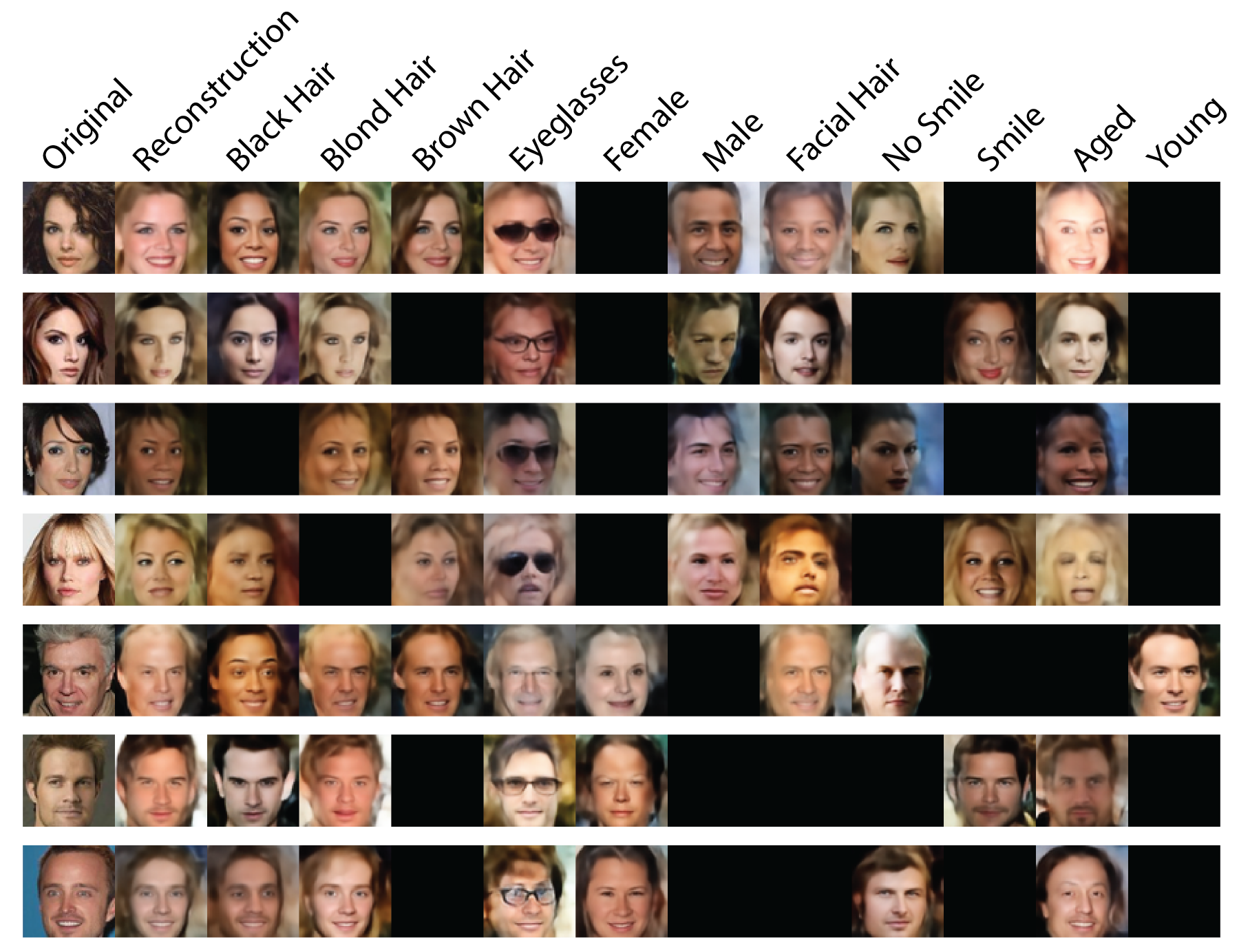}
\caption{Identity-distorting transformations with CGAN actor-critic. Without a penalty to encourage small moves in latent space, the actor maps the latent vectors of the original data points to generated images that have the correct attributes, but a different identity. Panels are black for attributes of the original image, as the procedure just returns the same image as the reconstruction.}
\label{fig:transform_cgan}
\end{figure}

Figure~\ref{fig:transform} demonstrates transformations applied to samples from the held-out evaluation dataset. Note that since the reconstructions are close to the original images, the transformed images also maintain much of their structure. This contrasts with supplemental Figure~\ref{fig:transform_cgan}, where a distance-penalty-free CGAN actor produces transformations that share attributes with the original but shift identity. We could preserve identity by introducing a distance penalty, but find that it is much easier to find the correct weighting of realism cost, attribute cost, and distance penalty through optimization, as each combination does not require retraining the network.

%%% ==============================================
\begin{figure}[t]
\centering
\includegraphics[width=0.8\textwidth]{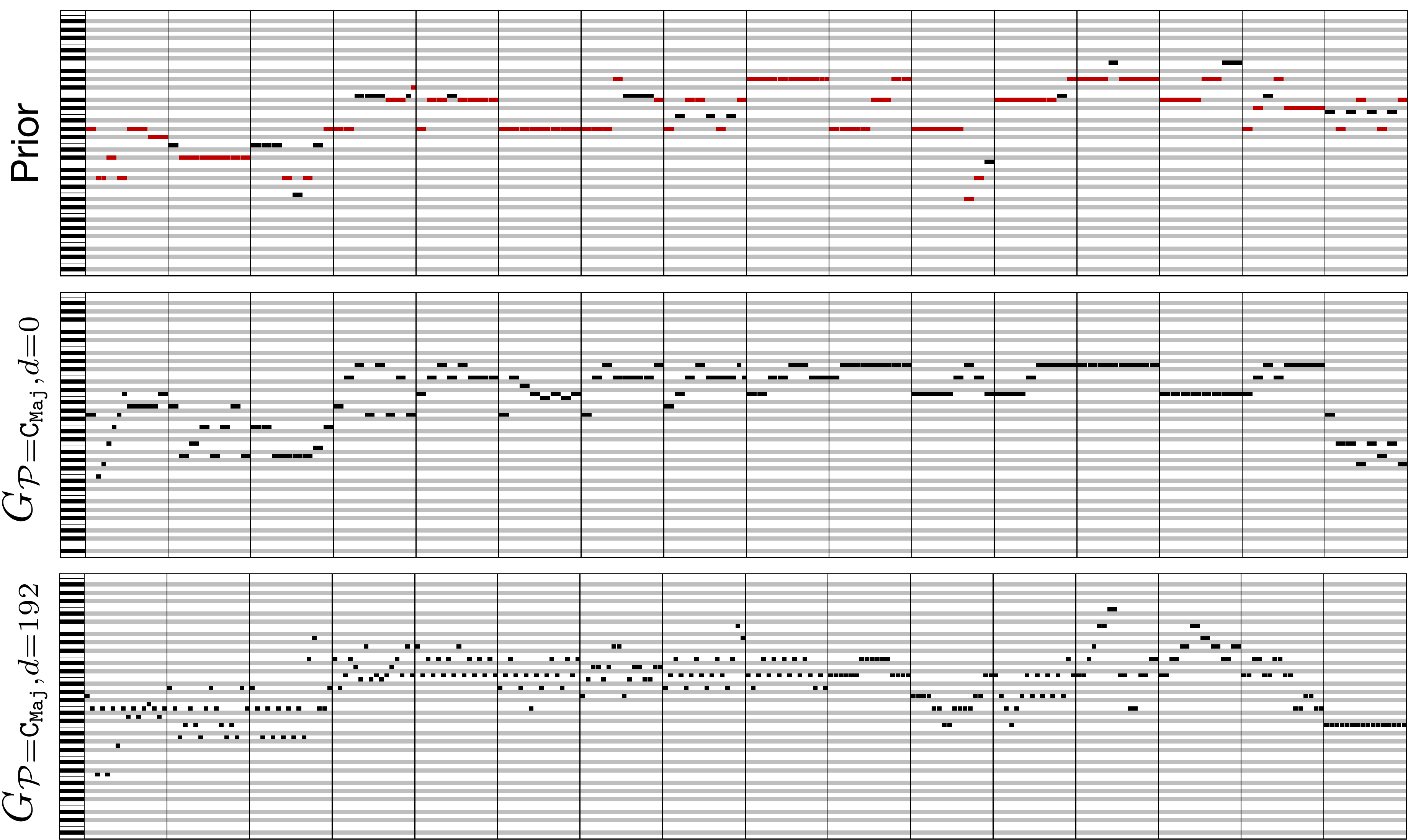}
\caption{Transformations from a prior sample for the Melody VAE model. In each 16-bar pianoroll, time is in the horizontal direction and pitch in the vertical direction. In the prior sample, notes falling outside of the C Major scale are shown in red. After transformation by $G_{\mathcal{P}=\CMaj, d=0}$, all sampled notes fall within the scale, without a significant change to note density. After transformation of the original $z$ by $G_{\mathcal{P}=\CMaj, d=192}$, all sampled notes lay within the scale and the density increases beyond 192. Synthesized audio of these samples can be heard at \url{https://goo.gl/ouULt9}.}
\label{fig:pianorolls}
\end{figure}
%%% ==============================================

%%% ==============================================
\section{Rule-based Constraints: Zero-shot Conditional Generation}
\label{sec:rules}

So far, we have assumed
%post-hoc
access to labeled data to train attribute classifiers. We can remove the need to provide labeled examples by leveraging the structure learned by our pre-trained model, using it to generate exemplars that are scored by a user-supplied reward function. If we constrain the reward function to be bounded, $c(x): \R^N \rightarrow [0, 1]$, the problem becomes very similar to previous GAN settings, but now the actor, $G$, and critic, $D$, are working together. $D$ aims to best approximate the true value of each latent state, $\E_{x \sim p(x\mid z)} c(x)$, and $G$ aims to shift samples from the prior to high-value states. The critic loss is the cross-entropy from $c(x)$, and the actor loss is the same as $\Loss_G$ in equation~\ref{eq:actor}, where we again have a distance penalty to promote diversity of outputs. 

% %
% \begin{equation}\begin{split}
% &\Loss_D(z) = \E_{z \sim q(z \mid x)}[\Loss_{c=1}(z)] + \E_{z \sim p(z)}[\Loss_{c=0}(z)] + \E_{z \sim G(p(z))}[\Loss_{c=0}(z)]\\
% &\Loss_G(z) = \E_{z \sim p(z)}[\log G(z)) + \lambda_\mathrm{dist} \Loss_{\mathrm{dist}}(G(z), z)]\\
% \end{split}\label{eq:rules}\end{equation}
% %

Note that the reward function and VAE decoder need not necessarily be differentiable, as the critic learns a value function to approximate the reward, which the actor uses for training. To highlight this, we demonstrate that the output of a recurrent VAE model can be constrained to satisfy hard-coded rule-based constraints. 

\begin{figure}[ht]
\includegraphics[width=\textwidth]{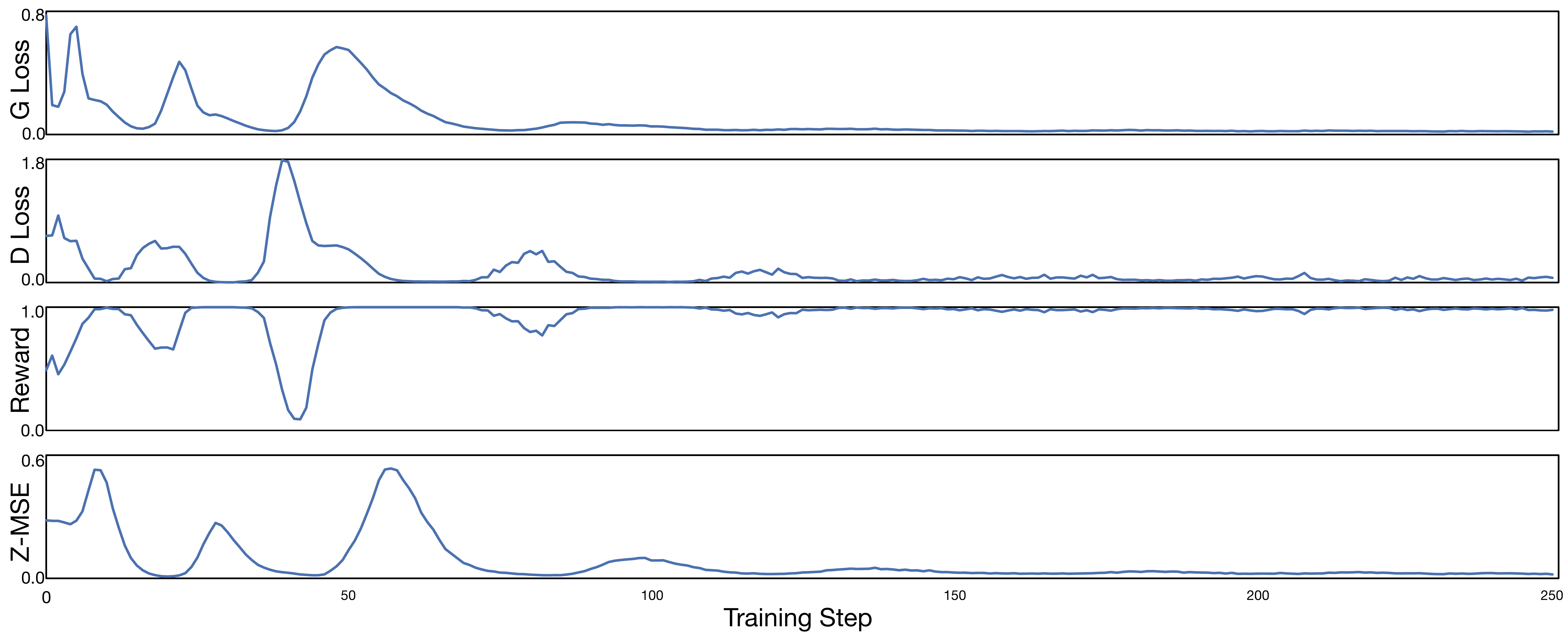}
\caption{Training curves for melody actor ($G$) and critic ($D$) pair for pitch class constraint $c_\mathrm{pitch}(m, \mathcal{P}=\CMaj)$.}
\label{fig:mel_training}
\end{figure}

We first train an LSTM VAE (details in the Appendix) on melodic fragments. Each melody, $m$, is represented as a sequence of categorical variables. In order to examine our ability to constrain the pitch classes and note density of the outputs, we define two reward functions, one that encourages notes from a set of pitches $\mathcal{P}$, and another for that encourages melodies to have at least $d$ notes:
\begin{equation}\begin{split}
&\textstyle c_\mathrm{pitch}(m, \mathcal{P}) = \sum_{p\in m}\mathbbm{1}(p \in \mathcal{P}) / |m| \qquad c_\mathrm{density}(m, d) = \mathrm{min}(1, |m| / d)
\end{split}\end{equation}
Figure~\ref{fig:pianorolls} gives an example of controlling the pitch class and note density of generated outputs, which is quantitatively supported by the results in Table~\ref{tab:musicrewards}. During training, the actor goes through several phases of exploration and exploitation, oscillating between expanding to find new modes with high reward and then contracting to find the nearest locations of those modes, eventually settling into high value states that require only small movements in the latent space (supplemental Figure~\ref{fig:mel_training}).

%%% ==============================================
\begin{table}[ht]
\centering
\begin{tabular}{ l | c c | c }
 \textbf{Actor} & $c_\mathrm{pitch}(m, \mathcal{P}=\CMaj)$ & $c_\mathrm{density}(m, d=192)$ & $z_\mse$ \\
 \hline
 Prior & 0.579 (0.43\%) & 0.417 (0.04\%) & - \\ 
 $G_{\mathcal{P}=\CMaj, d=0}$ & 0.991 (70.8\%) & 0.459 (0.01\%) & 0.015\\  
 $G_{\mathcal{P}=\CMaj, d=192}$ & 0.982 (62.4\%) & 0.985 (84.9\%) & 0.039   
\end{tabular}
\caption{Average rewards and constraint satisfaction rates (in parentheses) for unconditional (Prior) and conditional generation. Samples from the prior receive low rewards, on average, and near zero satisfaction rates from both the pitch class (C Major) and note density ($\geq192$ notes) constraints. After applying an actor optimized only for the C Major scale ($G_{\mathcal{P}=\CMaj, d=0}$), the pitch class constraint is fully satisfied 70.8\% of the time with only a minor effect on density. The average value close to 1 also indicates that when the constraint is not satisfied, it is typically off by only a few notes. Applying an actor function optimized for the C Major scale and high density ($G_{\mathcal{P}=\CMaj, d=192}$) causes both constraints to be satisfied at high rates, with a slightly larger shift in latent space.}
\label{tab:musicrewards}
\end{table}

\section{Related Work}

Conditional GANs \citep{Mirza2014} and VAEs \citep{Sohn2015} introduce conditioning
variables at training time. \citet{Sohn2015} allow these variables to affect the
distribution in latent $z$ space, but still require that $p(z\mid y)$
be a tractable distribution. \citet{Perarnau2016} use CGANs to adjust images,
but because CGANs cannot usually reconstruct arbitrary inputs accurately, they
must resort to image-space processing techniques to transfer effects to the original
input. \citet{white2016sampling} propose adding ``attribute vectors'' to samples from
$p(z)$ as a simple and effective heuristic to perform transformations,
which relies heavily on the linearity of the latent space.

Some recent work has focused on applying more expressive prior constraints to
VAEs \citep{rezende2014stochastic,Sonderby2016,Chen2016,Tomczak2017}. The prior that maximizes the
ELBO is $p^\star(z)=q(z)$ \citep{ELBOSurgery}; one can interpret our realism
constraint as trying to find an implicit distribution that is indistinguishable
from $q(z)$. Like the adversarial autoencoder of \citet{AAE}, our realism
constraint relies on a discriminative model, but instead of trying to force
$q(z)$ to equal some simple $p(z)$, we only weakly constrain $q(z)$ and then
use a classifier to ``clean up'' our results.

Like this work, the recently proposed
adversarially regularized autoencoder \citep{Junbo2017} uses adversarial training to generate
latent codes in a latent space discovered by an autoencoder; that work focuses
on unconditional generation.
\citet{Gomezbombarelli:2016} train classifiers in the latent space of a VAE to predict what latent variables map to molecules with various properties, and then use iterative gradient-based optimization in the latent space to find molecules that have a desired set of properties.
On molecule data, their procedure generates invalid molecules rarely enough that they can simply reject these samples, which are detected using off-the-shelf software. By contrast, the probability of generating realistic images under our pretrained VAE is astronomically small, and no simple criterion for detecting valid images exists.

\citet{Jaques2017} also use a classifier to constrain generation; they use a Deep Q-network as an auxiliary loss for training an LSTM. 
Closest to Section~\ref{sec:rules}, \citet{nguyen2016synthesizing,nguyen2016plug} generate very high quality conditional images by optimizing a sample from the latent space of a generative network to create an image that maximizes the class activations of a pretrained ImageNet classifier. Our work differs in that we learn an amortized generator/discriminator directly in the latent space and we achieve diversity through regularizing by the natural scale of the latent space rather than through a modified Langevin sampling algorithm.

% \textbf{}
% alphaGan \citep{alphaGAN}
% vamp \citep{wu2017vamp}
% attribute vector \citep{white2016sampling}
% cgan \citep{Mirza2014}
% gen adv text to image synthesis \citep{Reed2016}
% imporoved training of wassertsein gans \citep{Gulrajani2017}
% var Inf using Implicit Distributions \citep{Huszar2017}
% Invertible CGANS for image editing \citep{Perarnau2016}
% ALI \citep{Dumoulin2016}
% Adv. Autoencoders \citep{AAE} 
% Adv. Var. Bayes \citep{Mescheder2017} 
% Var. ? 
% IAF \citep{Kingma2016}
% note on eval. of generative models
% Sequence Tutor (for DQNN learning of rules). \citep{Jaques2017}
% Adversarial regularized autoencoders \citep{Junbo2017}

% \jcom{should probably also cite `Synthesizing the preferred inputs for neurons in
% neural networks via deep generator networks', which glues a generative model and a classifier together, to constrain the samples from the generator to satisfy (maximize) the class constraints \citep[][]{nguyen2016synthesizing}}

%%% ==============================================

% \section{Experiments}

%%% ==============================================

\section{Discussion and Future Work}
We have demonstrated a new approach to conditional generation by constraining the latent space of an unconditional generative model. This approach could be extended in a number of ways.

One possibility would be to plug in different architectures, including powerful autoregressive decoders or adversarial decoder costs, as we make no assumptions specific to independent likelihoods. While we have considered constraints based on implicit density estimation, we could also estimate the constrained distribution directly with an explicit autoregressive model or another variational autoencoder. The efficacy of autoregressive priors in VAEs is promising for this approach \citep{Kingma2016}. Conditional samples could then be obtained by ancestral sampling, and transformations by using gradient ascent to increase the likelihood under the model.
Active or semisupervised learning approaches could reduce the sample complexity of learning constraints.
Real-time constraint learning would also enable new applications; it might be fruitful to extend the reward approximation of Section~\ref{sec:rules} to incorporate user preferences as in \citep{christiano2017deep}.

\subsubsection*{Acknowledgments}
% NO ACKNOWLEDGMENTS FOR DOUBLE-BLIND SUBMISSION!
Many thanks to Jascha Sohl-Dickstein, Colin Raffel, and Doug Eck for their helpful brainstorming and encouragement. 

%acknowledgments, including those to funding agencies, go at the end of the paper.

% \bibliography{output.bbl}

% \bibliography{iclr2018_conference}

\begin{thebibliography}{33}
\providecommand{\natexlab}[1]{#1}
\providecommand{\url}[1]{\texttt{#1}}
\expandafter\ifx\csname urlstyle\endcsname\relax
  \providecommand{\doi}[1]{doi: #1}\else
  \providecommand{\doi}{doi: \begingroup \urlstyle{rm}\Url}\fi

\bibitem[Bernardo et~al.(2017)Bernardo, Zbyszyński, Fiebrink, and
  Grierson]{Bernardo2017}
Bernardo, Zbyszyński, Fiebrink, and Grierson.
\newblock {Interactive machine learning for end-user innovation}.
\newblock In \emph{Proceedings of the AAAI Symposium Series: Designing the User
  Experience of Machine Learning Systems}, 2017.
\newblock URL
  \url{http://research.gold.ac.uk/19767/1/BernardoZbyszynskiFiebrinkGrierson_UXML_2017.pdf}.

\bibitem[Bishop(2006)]{bishop2006}
Christopher~M Bishop.
\newblock Pattern recognition and machine learning (information science and
  statistics) springer-verlag new york.
\newblock \emph{Inc. Secaucus, NJ, USA}, 2006.

\bibitem[Brock et~al.(2016)Brock, Lim, Ritchie, and Weston]{Brock2016}
Andrew Brock, Theodore Lim, J.~M. Ritchie, and Nick Weston.
\newblock {Neural Photo Editing with Introspective Adversarial Networks}.
\newblock \emph{arXiv preprint}, 2016.
\newblock URL \url{https://arxiv.org/abs/1609.07093}.

\bibitem[Chen et~al.(2017)Chen, Kingma, Salimans, Duan, Dhariwal, Schulman,
  Sutskever, and Abbeel]{Chen2016}
Xi~Chen, Diederik~P. Kingma, Tim Salimans, Yan Duan, Prafulla Dhariwal, John
  Schulman, Ilya Sutskever, and Pieter Abbeel.
\newblock {Variational Lossy Autoencoder}.
\newblock In \emph{Proceedings of the International Conference on Learning
  Representations (ICLR)}, 2017.
\newblock URL \url{http://arxiv.org/abs/1611.02731}.

\bibitem[Christiano et~al.(2017)Christiano, Leike, Brown, Martic, Legg, and
  Amodei]{christiano2017deep}
Paul Christiano, Jan Leike, Tom~B Brown, Miljan Martic, Shane Legg, and Dario
  Amodei.
\newblock Deep reinforcement learning from human preferences.
\newblock \emph{arXiv preprint}, 2017.
\newblock URL \url{https://arxiv.org/abs/1706.03741}.

\bibitem[Donahue et~al.(2016)Donahue, Kr{\"a}henb{\"u}hl, and
  Darrell]{donahue2016adversarial}
Jeff Donahue, Philipp Kr{\"a}henb{\"u}hl, and Trevor Darrell.
\newblock Adversarial feature learning.
\newblock \emph{arXiv preprint arXiv:1605.09782}, 2016.

\bibitem[Dumoulin et~al.(2016)Dumoulin, Belghazi, Poole, Mastropietro, Lamb,
  Arjovsky, and Courville]{Dumoulin2016}
Vincent Dumoulin, Ishmael Belghazi, Ben Poole, Olivier Mastropietro, Alex Lamb,
  Martin Arjovsky, and Aaron Courville.
\newblock {Adversarially Learned Inference}.
\newblock In \emph{Proceedings of the International Conference on Learning
  Representations (ICLR)}, 2016.
\newblock URL \url{https://arxiv.org/abs/1606.00704}.

\bibitem[{G{\'o}mez-Bombarelli} et~al.(2016){G{\'o}mez-Bombarelli}, {Wei},
  {Duvenaud}, {Hern{\'a}ndez-Lobato}, {S{\'a}nchez-Lengeling}, {Sheberla},
  {Aguilera-Iparraguirre}, {Hirzel}, {Adams}, and
  {Aspuru-Guzik}]{Gomezbombarelli:2016}
R.~{G{\'o}mez-Bombarelli}, J.~N. {Wei}, D.~{Duvenaud}, J.~M.
  {Hern{\'a}ndez-Lobato}, B.~{S{\'a}nchez-Lengeling}, D.~{Sheberla},
  J.~{Aguilera-Iparraguirre}, T.~D. {Hirzel}, R.~P. {Adams}, and
  A.~{Aspuru-Guzik}.
\newblock {Automatic chemical design using a data-driven continuous
  representation of molecules}.
\newblock \emph{ArXiv e-prints}, October 2016.

\bibitem[Goodfellow et~al.(2014)Goodfellow, Pouget-Abadie, Mirza, Xu,
  Warde-Farley, Ozair, Courville, and Bengio]{gan}
Ian Goodfellow, Jean Pouget-Abadie, Mehdi Mirza, Bing Xu, David Warde-Farley,
  Sherjil Ozair, Aaron Courville, and Yoshua Bengio.
\newblock Generative adversarial nets.
\newblock In \emph{Advances in Neural Information Processing Systems (NIPS)},
  2014.
\newblock URL
  \url{http://papers.nips.cc/paper/5423-generative-adversarial-nets.pdf}.

\bibitem[Gulrajani et~al.(2017)Gulrajani, Ahmed, Arjovsky, Dumoulin, and
  Courville]{Gulrajani2017}
Ishaan Gulrajani, Faruk Ahmed, Martin Arjovsky, Vincent Dumoulin, and Aaron
  Courville.
\newblock {Improved Training of Wasserstein GANs}.
\newblock \emph{arXiv preprint}, 2017.
\newblock URL \url{http://arxiv.org/abs/1704.00028}.

\bibitem[Hoffman \& Johnson(2016)Hoffman and Johnson]{ELBOSurgery}
Matthew~D. Hoffman and Matthew~J. Johnson.
\newblock {ELBO surgery: yet another way to carve up the variational evidence
  lower bound}.
\newblock In \emph{Workshop in Advances in Approximate Bayesian Inference,
  NIPS}, 2016.
\newblock URL
  \url{http://approximateinference.org/accepted/HoffmanJohnson2016.pdf}.

\bibitem[Jaques et~al.(2017)Jaques, Gu, Bahdanau, Hernández-Lobato, Turner,
  and Eck]{Jaques2017}
Natasha Jaques, Shixiang Gu, Dzmitry Bahdanau, José~Miguel Hernández-Lobato,
  Richard~E. Turner, and Douglas Eck.
\newblock Sequence tutor: Conservative fine-tuning of sequence generation
  models with kl-control.
\newblock In \emph{Proceedings of the International Conference on Learning
  Representations (ICLR)}, 2017.
\newblock URL \url{https://arxiv.org/abs/1611.02796}.

\bibitem[Johnson et~al.(2016)Johnson, Alahi, and
  Fei-Fei]{johnson2016perceptual}
Justin Johnson, Alexandre Alahi, and Li~Fei-Fei.
\newblock Perceptual losses for real-time style transfer and super-resolution.
\newblock In \emph{European Conference on Computer Vision}, pp.\  694--711.
  Springer, 2016.

\bibitem[Junbo et~al.(2017)Junbo, Zhao, Kim, Zhang, Rush, and LeCun]{Junbo2017}
Junbo, Zhao, Yoon Kim, Kelly Zhang, Alexander~M. Rush, and Yann LeCun.
\newblock {Adversarially Regularized Autoencoders for Generating Discrete
  Structures}.
\newblock \emph{arXiv preprint}, 2017.
\newblock URL \url{http://arxiv.org/abs/1706.04223}.

\bibitem[Kingma \& Ba(2015)Kingma and Ba]{kingma2014adam}
Diederik~P. Kingma and Jimmy Ba.
\newblock Adam: {A} method for stochastic optimization.
\newblock In \emph{Proceedings of the International Conference on Learning
  Representations (ICLR)}, 2015.
\newblock URL \url{http://arxiv.org/abs/1412.6980}.

\bibitem[Kingma \& Welling(2013)Kingma and Welling]{Kingma2013}
Diederik~P. Kingma and Max Welling.
\newblock Auto-encoding variational bayes.
\newblock In \emph{Proceedings of the International Conference on Learning
  Representations (ICLR)}, 2013.
\newblock URL \url{http://arxiv.org/abs/1312.6114}.

\bibitem[Kingma et~al.(2016)Kingma, Salimans, Jozefowicz, Chen, Sutskever, and
  Welling]{Kingma2016}
Diederik~P. Kingma, Tim Salimans, Rafal Jozefowicz, Xi~Chen, Ilya Sutskever,
  and Max Welling.
\newblock {Improving Variational Inference with Inverse Autoregressive Flow}.
\newblock In \emph{Advances in Neural Information Processing Systems (NIPS)},
  2016.
\newblock URL \url{http://arxiv.org/abs/1606.04934}.

\bibitem[LeCun et~al.(2010)LeCun, Cortes, and Burges]{lecun2010mnist}
Yann LeCun, Corinna Cortes, and Christopher~JC Burges.
\newblock Mnist handwritten digit database. at\&t labs, 2010.

\bibitem[Li \& Wand(2016)Li and Wand]{li2016precomputed}
Chuan Li and Michael Wand.
\newblock Precomputed real-time texture synthesis with markovian generative
  adversarial networks.
\newblock In \emph{European Conference on Computer Vision}, pp.\  702--716.
  Springer, 2016.

\bibitem[Liu et~al.(2015)Liu, Luo, Wang, and Tang]{liu2015faceattributes}
Ziwei Liu, Ping Luo, Xiaogang Wang, and Xiaoou Tang.
\newblock Deep learning face attributes in the wild.
\newblock In \emph{Proceedings of International Conference on Computer Vision
  (ICCV)}, 2015.
\newblock URL \url{https://arxiv.org/abs/1411.7766}.

\bibitem[Makhzani et~al.(2016)Makhzani, Shlens, Jaitly, and Goodfellow]{AAE}
Alireza Makhzani, Jonathon Shlens, Navdeep Jaitly, and Ian Goodfellow.
\newblock Adversarial autoencoders.
\newblock In \emph{Proceedings of the International Conference on Learning
  Representations (ICLR)}, 2016.
\newblock URL \url{http://arxiv.org/abs/1511.05644}.

\bibitem[Mirza \& Osindero(2014)Mirza and Osindero]{Mirza2014}
Mehdi Mirza and Simon Osindero.
\newblock {Conditional Generative Adversarial Nets}.
\newblock \emph{arXiv preprint}, 2014.
\newblock URL \url{http://arxiv.org/abs/1411.1784}.

\bibitem[Nguyen et~al.(2016{\natexlab{a}})Nguyen, Dosovitskiy, Yosinski, Brox,
  and Clune]{nguyen2016synthesizing}
Anh Nguyen, Alexey Dosovitskiy, Jason Yosinski, Thomas Brox, and Jeff Clune.
\newblock Synthesizing the preferred inputs for neurons in neural networks via
  deep generator networks.
\newblock In \emph{Advances in Neural Information Processing Systems (NIPS)},
  2016{\natexlab{a}}.
\newblock URL \url{https://arxiv.org/abs/1605.09304}.

\bibitem[Nguyen et~al.(2016{\natexlab{b}})Nguyen, Yosinski, Bengio,
  Dosovitskiy, and Clune]{nguyen2016plug}
Anh Nguyen, Jason Yosinski, Yoshua Bengio, Alexey Dosovitskiy, and Jeff Clune.
\newblock Plug \& play generative networks: Conditional iterative generation of
  images in latent space.
\newblock \emph{arXiv preprint arXiv:1612.00005}, 2016{\natexlab{b}}.

\bibitem[Perarnau et~al.(2016)Perarnau, van~de Weijer, Raducanu, and
  {\'{A}}lvarez]{Perarnau2016}
Guim Perarnau, Joost van~de Weijer, Bogdan Raducanu, and Jose~M. {\'{A}}lvarez.
\newblock {Invertible Conditional GANs for image editing}.
\newblock In \emph{Workshop on Adversarial Training, NIPS}, 2016.
\newblock URL \url{http://arxiv.org/abs/1611.06355
  http://www.cvc.uab.es/LAMP/wp-content/uploads/Projects/pdfs/presentationNIPS.pdf}.

\bibitem[Radford et~al.(2015)Radford, Metz, and
  Chintala]{DBLP:journals/corr/RadfordMC15}
Alec Radford, Luke Metz, and Soumith Chintala.
\newblock Unsupervised representation learning with deep convolutional
  generative adversarial networks.
\newblock \emph{CoRR}, abs/1511.06434, 2015.
\newblock URL \url{http://arxiv.org/abs/1511.06434}.

\bibitem[Rezende et~al.(2014)Rezende, Mohamed, and
  Wierstra]{rezende2014stochastic}
Danilo~Jimenez Rezende, Shakir Mohamed, and Daan Wierstra.
\newblock Stochastic backpropagation and approximate inference in deep
  generative models.
\newblock \emph{arXiv preprint arXiv:1401.4082}, 2014.

\bibitem[Salimans et~al.(2016)Salimans, Goodfellow, Zaremba, Cheung, Radford,
  Chen, and Chen]{Salimans2016}
Tim Salimans, Ian Goodfellow, Wojciech Zaremba, Vicki Cheung, Alec Radford,
  Xi~Chen, and Xi~Chen.
\newblock Improved techniques for training gans.
\newblock In \emph{Advances in Neural Information Processing Systems 29}, 2016.
\newblock URL
  \url{http://papers.nips.cc/paper/6125-improved-techniques-for-training-gans.pdf}.

\bibitem[Sohn et~al.(2015)Sohn, Lee, and Yan]{Sohn2015}
Kihyuk Sohn, Honglak Lee, and Xinchen Yan.
\newblock Learning structured output representation using deep conditional
  generative models.
\newblock In \emph{Advances in Neural Information Processing Systems (NIPS)},
  2015.
\newblock URL
  \url{http://papers.nips.cc/paper/5775-learning-structured-output-representation-using-deep-conditional-generative-models.pdf}.

\bibitem[S{\o}nderby et~al.(2016)S{\o}nderby, Raiko, Maal{\o}e, S{\o}nderby,
  and Winther]{Sonderby2016}
Casper~Kaae S{\o}nderby, Tapani Raiko, Lars Maal{\o}e, S{\o}ren~Kaae
  S{\o}nderby, and Ole Winther.
\newblock Ladder variational autoencoders.
\newblock In \emph{Advances in Neural Information Processing Systems}, pp.\
  3738--3746, 2016.

\bibitem[Tomczak \& Welling(2017)Tomczak and Welling]{Tomczak2017}
Jakub~M. Tomczak and Max Welling.
\newblock {VAE} with a {VampPrior}.
\newblock \emph{CoRR}, abs/1705.07120, 2017.
\newblock URL \url{http://arxiv.org/abs/1705.07120}.

\bibitem[Ulyanov et~al.(2016)Ulyanov, Lebedev, Vedaldi, and
  Lempitsky]{Ulyanov2016}
Dmitry Ulyanov, Vadim Lebedev, Andrea Vedaldi, and Victor~S. Lempitsky.
\newblock Texture networks: Feed-forward synthesis of textures and stylized
  images.
\newblock In \emph{Proceedings of the 33rd International Conference on Machine
  Learning (ICML)}, 2016.
\newblock URL \url{http://arxiv.org/abs/1603.03417}.

\bibitem[White(2016)]{white2016sampling}
Tom White.
\newblock Sampling generative networks: Notes on a few effective techniques.
\newblock \emph{arXiv preprint}, 2016.
\newblock URL \url{https://arxiv.org/abs/1609.04468}.

\end{thebibliography}
\bibliographystyle{iclr2018_conference}

%%% ==============================================
\newpage
\section{Appendix}

\subsection{Experimental Details}
For images, we use the MNIST digits dataset \citep[][]{lecun2010mnist} and the Large-scale CelebFaces Attributes (CelebA) dataset \citep{liu2015faceattributes}. MNIST images are $28\times28$ pixels and greyscale scaled to [0, 1]. For attributes, we use the number class label of each digit. CelebA images are center-cropped to $128\times128$ pixels and then downsampled to $64\times64$ RGB pixels and scaled to [0, 1]. We find that many of the attribute labels are not strongly correlated with changes in the images, so we narrow the original 40 attributes to the 10 most visually salient: blond hair, black hair, brown hair, bald, eyeglasses, facial hair, hat, smiling, gender, and age.

For melodies, we scraped the web to collect over 1.5 million publicly available MIDI files. We then extracted 16-bar melodies by sliding a window with a single bar stride over each non-percussion instrument with a $\frac{4}{4}$ time signature, keeping only the note with the highest pitch when multiple overlap. This produced over 3 million unique melodies. We represent each melody as a sequence of 256 (16 per bar) categorical variables taking one of 130 discrete states at each sixteenth note: 128 note-on pitches, a hold state, and a rest state.

\subsection{Model Architectures}
\label{sec:arch}
 All encoders, decoders, and classifiers are trained with the Adam optimizer \citep[][]{kingma2014adam}, with learning rate $=$ 3e-4, $\beta_1 =$ 0.9, and $\beta_2 =$ 0.999. 

To train $D_{real}(z)$, $D_{attr}(z)$ and $G(z)$ we follow the training procedure of \cite{Gulrajani2017}, applying a gradient penalty of 10, training $D$ and $G$ in a 10:1 step ratio, and use the Adam optimizer with learning rate $=$ 3e-4, $\beta_1 =$ 0.0, and $\beta_2 =$ 0.9.  While not necessary to converge, we find it improves the stability of optimization. We do not apply any of the other tricks of GAN training such as batch normalization, minibatch discrimination, or one-sided label smoothing \citep{DBLP:journals/corr/RadfordMC15, Salimans2016}. As samples from $p(z)$ are easier to discriminate than samples from $G(p(z))$, we train $D$ by sampling from $p(z)$ at a rate 10 times less than $G(p(z))$. For actors with inner-loop optimization, $G_{\mathrm{opt}}$, 100 iterations of Adam are used with with learning rate $=$ 1e-1, $\beta_1 =$ 0.9, and $\beta_2 =$ 0.999.

\subsubsection{MNIST Feed-forward VAE}
To model the MNIST data, we use a deep feed-forward neural network (Figure~\ref{fig:vaes}a).

The encoder is a series of 3 linear layers with 1024 outputs, each followed by a ReLU, after which an additional linear layer is used to produce 2048 outputs. Half of the outputs are used as the $\mu$ and the softplus of the other half are used as the $\sigma$ to parameterize a 1024-dimension multivariate Gaussian distribution with a diagonal covariance matrix for $z$. 

The decoder is a series of 3 linear layers with 1024 outputs, each followed by a ReLU, after which an additional linear layer is used to produce 28x28 outputs. These outputs are then passed through a sigmoid to generate the output image.

\subsubsection{CelebA Convolutional VAE}
To model the CelebA data, we use a deep convolutional neural network (Figure~\ref{fig:vaes}b).

The encoder is a series of 4 2D convolutional layers, each followed by a ReLU. The convolution kernels are of size $3\times3$, $3\times3$, $5\times5$, and $5\times5$, with 2048, 1024, 512, and 256 output channels, respectively. All convolutional layers have a stride of 2. After the final ReLU, a linear layer is used to produce 2048 outputs. Half of the outputs are used as the $\mu$ and the softplus of the other half are used as the $\sigma$ to parameterize a 1024-dimension multivariate Gaussian distribution with a diagonal covariance matrix for $z$. 

The decoder passes the $z$ through a 4x4x2048 linear layer, and then a series of 4 2D transposed convolutional layers, all but the last of which are followed by a ReLU. The deconvolution kernels are of size $5\times5$, $5\times5$, $3\times3$, and $3\times3$, with 1024, 512, 256, and 3 output channels, respectively. All deconvolution layers have a stride of 2. The output from the final deconvolution is passed through a sigmoid to generate the output image.

The classifier that is trained to predict labels from images are identical to the VAE encoders except that they end with a sigmoid cross-entropy loss.

\subsubsection{Melody Sequence VAE}
\begin{figure}[ht]
\includegraphics[width=\textwidth]{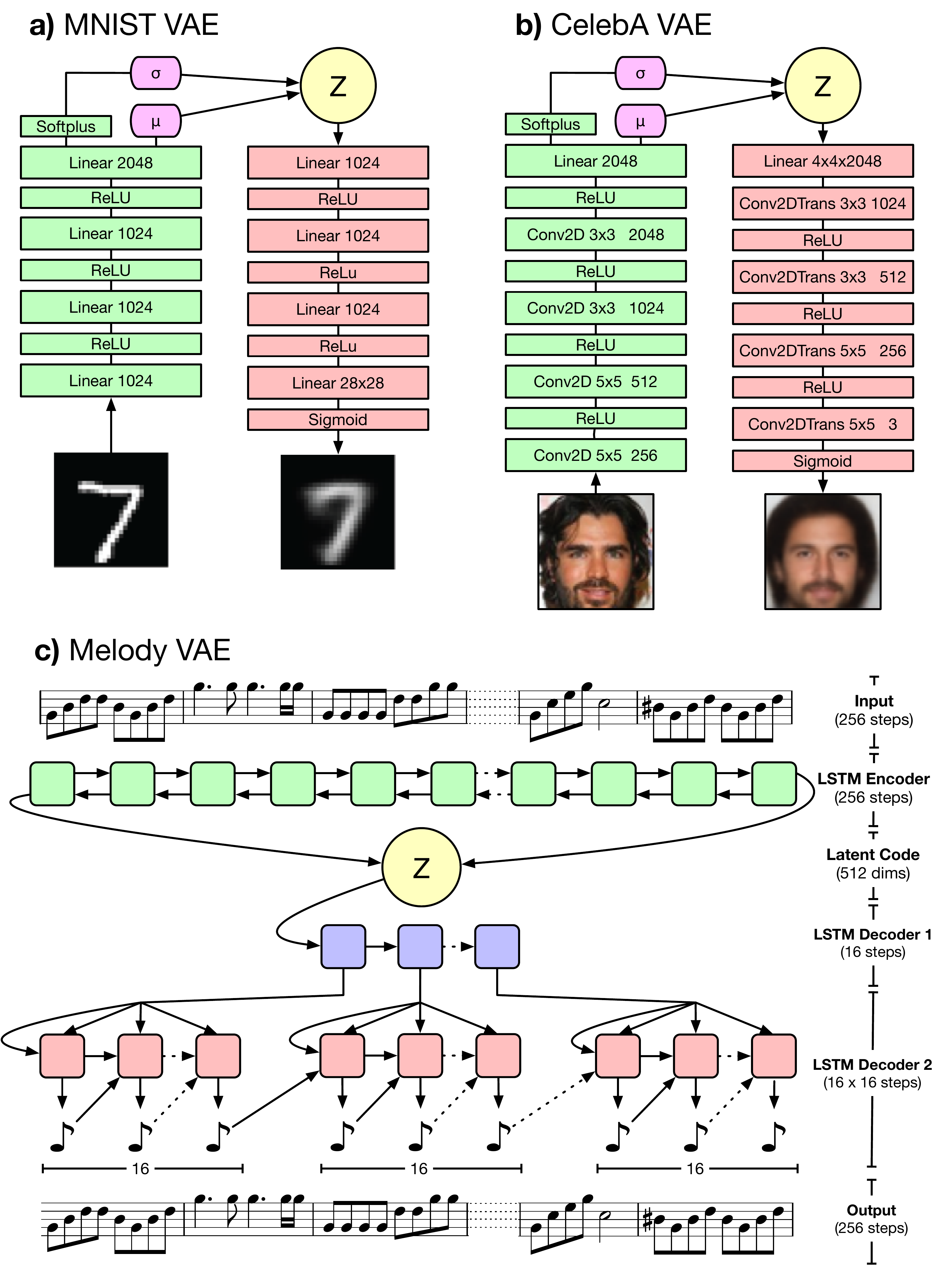}
\caption{Architectures for the (a) feed-forward MNIST, (b) convolutional CelebA, and (c) hierarchical LSTM melody VAEs. In (b), all convolutions have a stride of 2. In (c), LSTM cells shown in the same color share weights and linear layers between levels are omitted.}
\label{fig:vaes}
\end{figure}

Music is fundamentally sequential, so we use an LSTM-based sequence VAE for modelling monophonic melodies (Figure~\ref{fig:vaes}c).

The encoder is made up of a single-layer bidirectional LSTM, with 2048 units per cell. The final output in each direction is concatenated and passed through a linear layer to produce 1024 outputs. Half of the outputs are used as the $\mu$ and the softplus of the other half are used as a $\sigma$ to parameterize a 512-dimension multivariate Gaussian distribution with a diagonal covariance matrix for $z$. 

Since musical sequences often have structure at the bar level, we use a hierarchical decoder to model long melodies. First, the $z$ goes through a linear layer to initialize the state of a 2-layer LSTM with 1024 units per layer, which outputs 16 embeddings of size 512 each, one per bar. Each of these embeddings are passed through a linear layer to produce 16 initial states for another 2-layer LSTM with 1024 units per layer. This bar-level LSTM autoregressively produces individual sixteenth note events, passing its output through a linear layer and softmax to create a distribution over the 130 classes. This categorical distribution is used to compute a cross-entropy loss during training or samples at inference time. In addition to generating the initial state at the start of each bar, the embedding for the current bar is concatenated with the previous output as the input at each time step.

\subsubsection{Actor Feed-forward Network}
For $G(z)$, we use a deep feed-forward neural network (Figure~\ref{fig:actor_critic}a) in all of our experiments.

The network is a series of 4 linear layers with 2048 outputs, each followed by a ReLU, after which an additional linear layer is used to produce $2 * dim(z)$ outputs. Half of the outputs are used as the $\delta z$ and the sigmoid of the other half are used as $gates$. The transformed $z'$ is the computed as $(1-gates) * z + gates * \delta z$. This aids in training as the network only has to then predict shifts in $z$.

When conditioning on attribute labels, $y$, to compute $G(z, y)$, the labels are passed through a linear layer producing 2048 outputs which are concatenated with $z$ as the model input.

\subsubsection{Critic Feed-forward Network}
\begin{figure}[ht]
\centering
\includegraphics[width=0.5\textwidth]{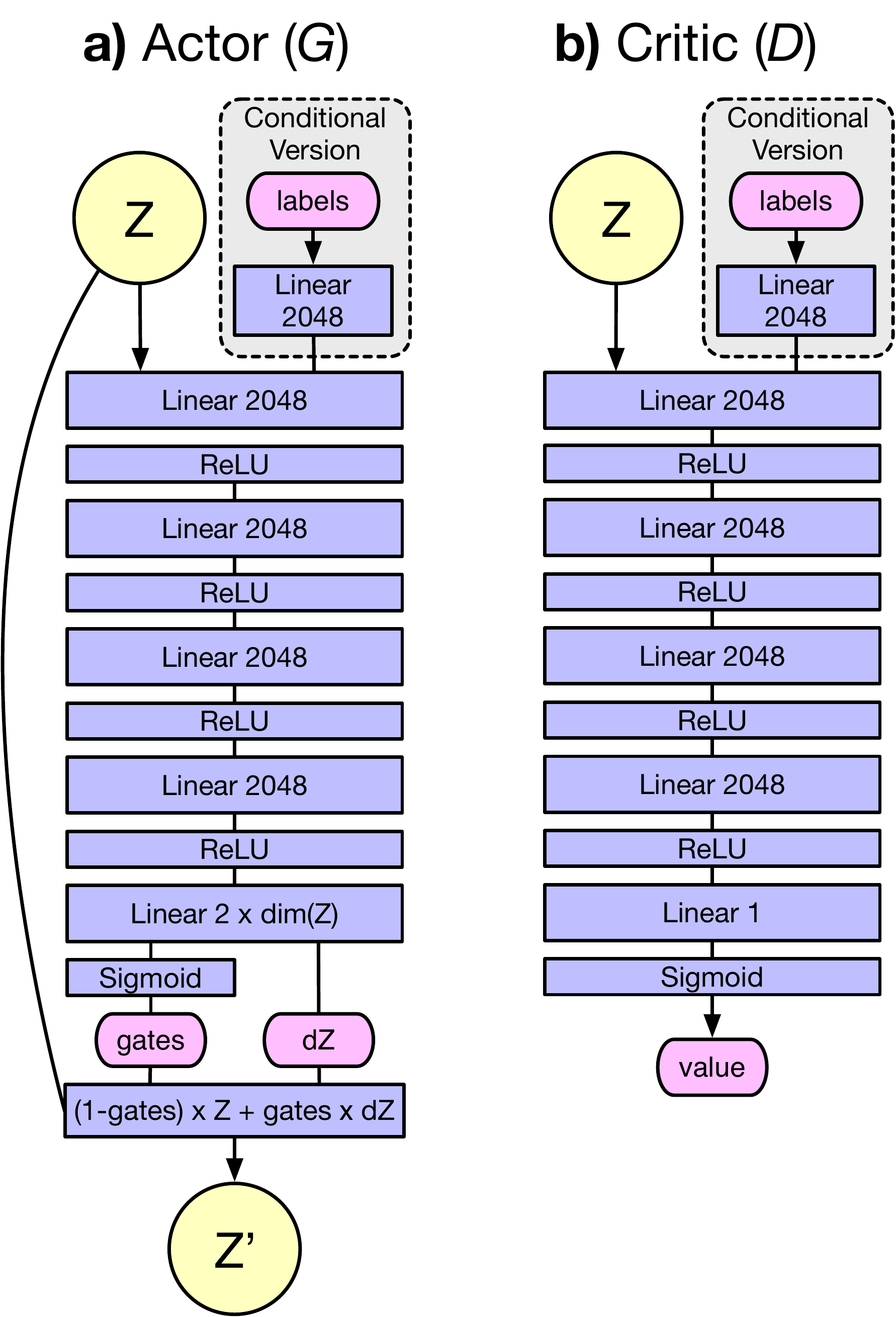}
\caption{Architecture for the (a) actors and (b) critics used in all experiments.}
\label{fig:actor_critic}
\end{figure}

For $D(z)$, we use a deep feed-forward neural network (Figure~\ref{fig:actor_critic}b) in all of our experiments.

The network is a series of 4 linear layers with 2048 outputs, each followed by a ReLU, after which an additional linear layer is used to produce a single output. This output is passed through a sigmoid to compute $D(z)$.

When conditioning on attribute labels, $y$, to compute $D(z, y)$, the labels are passed through a linear layer producing 2048 outputs which are concatenated with $z$ as the model input.
%%% ==============================================

\newpage

\subsection{Supplemental Figures}

\begin{figure}[ht]
\includegraphics[width=\textwidth]{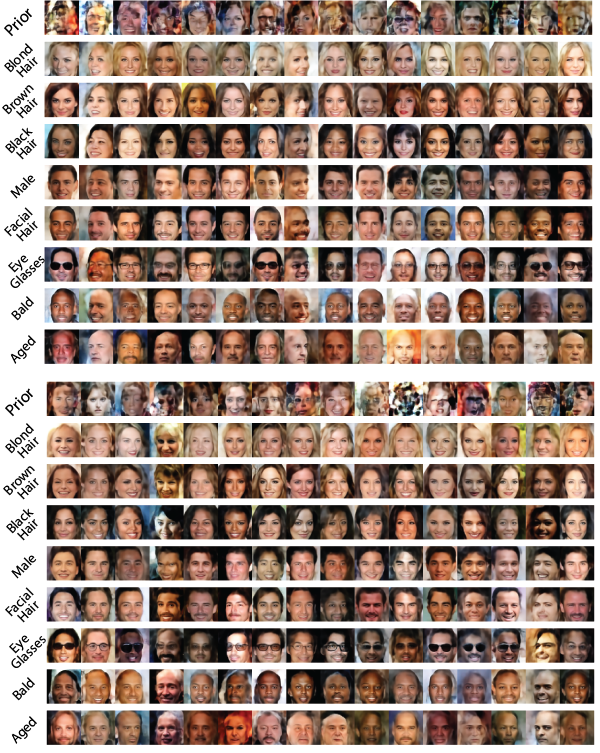}
\caption{Additional generated CelebA faces by $G_{\mathrm{CGAN}}$ with $\lambda_\mathrm{dist} = 0$. Full attribute labels are given in supplementary Table~\ref{tab:attrlist}}
\label{fig:suppcgan}
\end{figure}

\newpage

\begin{figure}[ht]
\includegraphics[width=\textwidth]{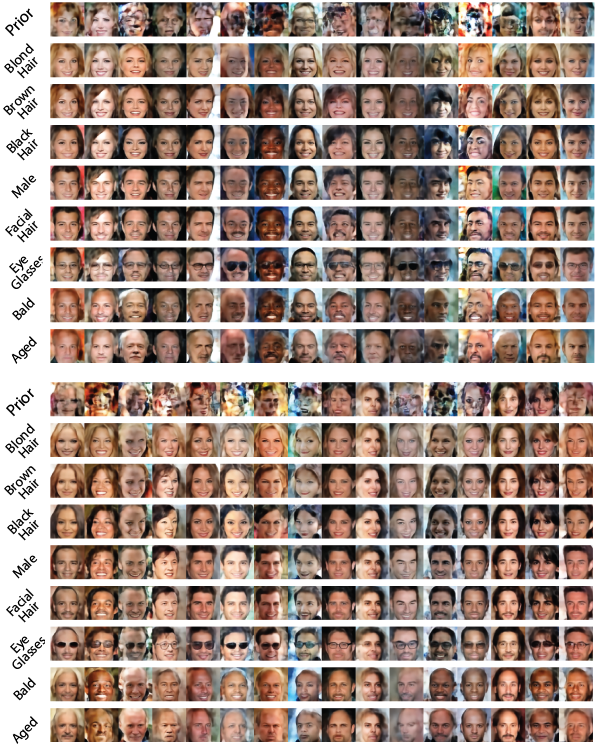}
\caption{Additional generated CelebA faces by $G_{\mathrm{CGAN}}$ with $\lambda_\mathrm{dist} = 0.1$. Full attribute labels are given in supplementary Table~\ref{tab:attrlist}}
\label{fig:suppcgandist}
\end{figure}

\newpage

\newpage

% \begin{figure}[ht]
% \includegraphics[width=\textwidth]{LatentConstraints_ICLR2018_Figs_ContourSupplemental.png}
% \caption{Latent dimensions with smaller $\bar{\sigma_z}$ are more important for reconstructions. Countours of the realism constraint in latent space. The two dimensions with smallest $\bar{\sigma_z}$ are on the left, and largest $\bar{\sigma_z}$ are on the right.}
% \label{fig:contour_supplemental}
% \end{figure}

\newpage

%%% ==============================================
\begin{table}[ht]
\centering
\begin{tabular}{ l | c c c c c c c c c c }
  \textbf{Figure}&      & Black & Blond & Brown & Eye- & & & & &\\
  \textbf{Label} & Bald & Hair & Hair & Hair & glasses & Male & Beard & Smiling & Hat & Young\\
 \hline
 Blond Hair &  0 & 0 & 1 & 0 & 0 & 0 & 0 & 1 & 0 & 1\\
 Brown Hair &  0 & 0 & 0 & 1 & 0 & 0 & 0 & 1 & 0 & 1\\
 Black Hair &  0 & 1 & 0 & 0 & 0 & 0 & 0 & 1 & 0 & 1\\
 Male &        0 & 1 & 0 & 0 & 0 & 1 & 0 & 1 & 0 & 1\\
 Facial Hair & 0 & 1 & 0 & 0 & 0 & 1 & 1 & 1 & 0 & 1\\
 Eyeglasses &  0 & 1 & 0 & 0 & 1 & 1 & 1 & 1 & 0 & 1\\
 Bald &        1 & 0 & 0 & 0 & 0 & 1 & 1 & 1 & 0 & 1\\
 Aged &        1 & 0 & 0 & 0 & 0 & 1 & 1 & 0 & 0 & 0\\
\end{tabular}
\caption{Complete list of attributes for label names in Figures~\ref{fig:cgan},~\ref{fig:suppcgan}, and ~\ref{fig:suppcgandist}}.
\label{tab:attrlist}
\end{table}
%%% ==============================================

%%% ==============================================
\begin{table}[ht]
\centering
\begin{tabular}{ l | c c c }
 $\sigma_x^2$ & LL & KL & ELBO \\
 \hline
 1    & -11360 & 30   & -11390 \\
 1e-1 & -11325 & 150  & -11475 \\
 \textbf{1e-2} &  15680 & 600  &  \textbf{15080} \\
 1e-3 &  16090 & 1950 &  14140 \\
 1e-4 &  16150 & 3650 &  12500 \\
 \end{tabular}
\caption{Selection of $\sigma_x=0.1$ for the CelebA VAEs by ELBO maximization. All results are given in Nats.}
\label{tab:ELBO}
\end{table}
%%% ==============================================

\newpage 

\begin{figure}[ht]
\includegraphics[width=\textwidth]{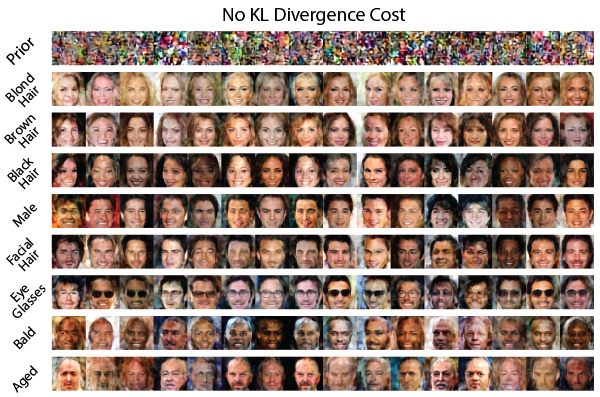}
\caption{Latent constraints applied to a vanilla autoencoder with no latent prior. Samples are similar quality to VAEs with $\sigma_x=0.1$, but with less diversity and more high-frequency visual artifacts. Full attribute labels are given in supplementary Table~\ref{tab:attrlist}.}
\label{fig:nokl}
\end{figure}

\newpage 

\begin{figure}[ht]
\includegraphics[width=\textwidth]{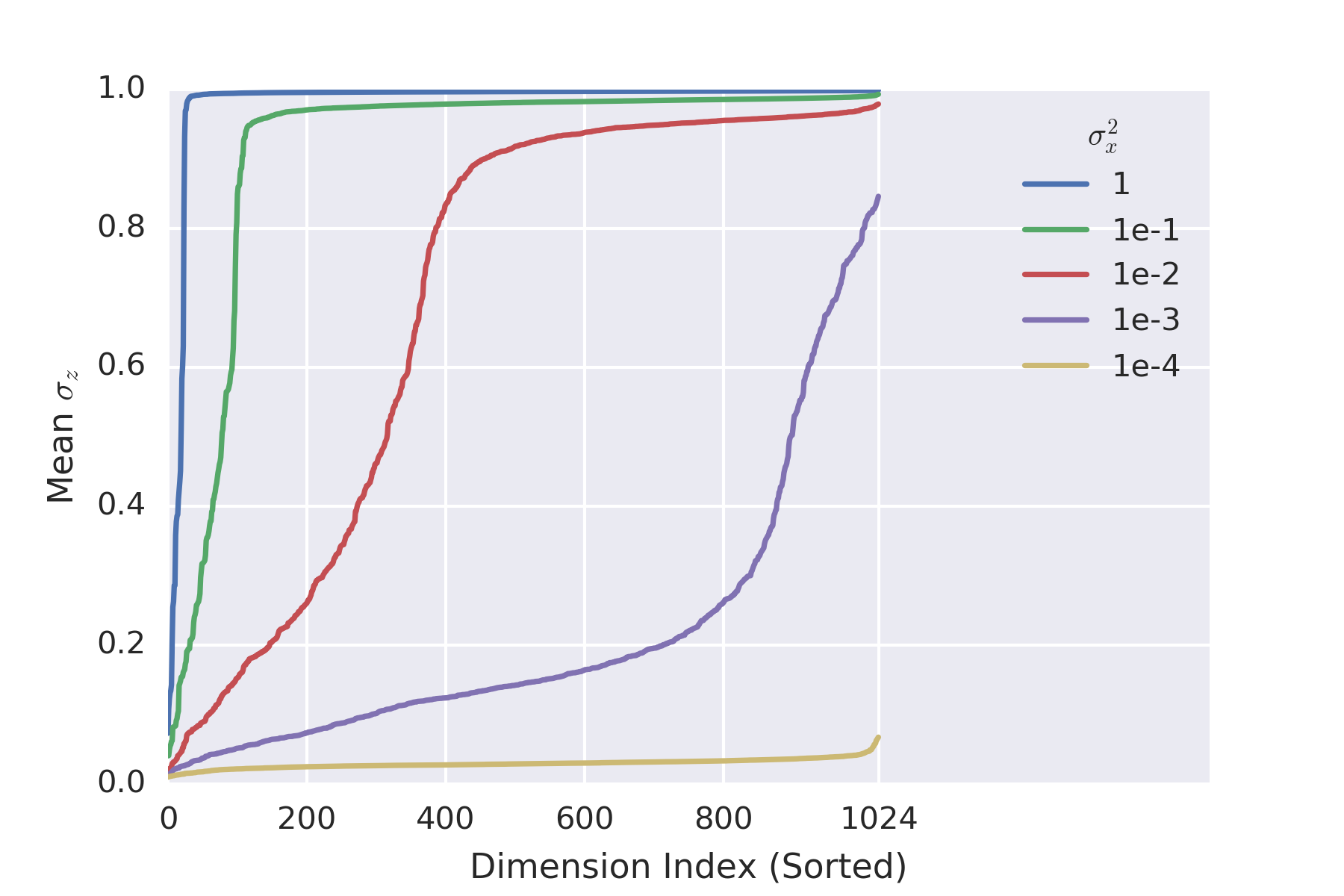}
\caption{Smaller decoder standard deviations, $\sigma_x$, lead to lower-variance posteriors, $\sigma_z(x)$ of the encoder $q(z\mid x)$, averaged over the training set per a dimension. The x-axis is sorted from lowest to highest variance. Tighter posteriors correspond to more utilization of the latent dimension, and we scale our distance regularization the square inverse on a per-dimension basis.}
\label{fig:sigmamean}
\end{figure}

\end{document}